\documentclass{article} 
\usepackage{iclr2027_conference,times}

\usepackage{amsmath,amsfonts,bm}

\def\eqref#1{equation~\ref{#1}}

\def\1{\bm{1}}

\DeclareMathAlphabet{\mathsfit}{\encodingdefault}{\sfdefault}{m}{sl}
\SetMathAlphabet{\mathsfit}{bold}{\encodingdefault}{\sfdefault}{bx}{n}

\usepackage{hyperref}
\usepackage{url}
\usepackage{graphicx}
\usepackage{xcolor}

\title{PertMind: Eliciting Emergent Biological Reasoning in LLM via Reinforcement Learning on Cellular Perturbation Data}

\author{%
\textbf{Zhenchao Tang}\textsuperscript{1,\textdagger}
\quad \textbf{Xiaogang Xu}\textsuperscript{1,\textdagger,*}
\quad \textbf{Jiafei Wu}\textsuperscript{1}
\quad \textbf{Jiahui Guan}\textsuperscript{2}
\quad \textbf{Bo Li}\textsuperscript{3}
\quad \textbf{Tianxu Lv}\textsuperscript{4}
\\[0.25em]
\textbf{Jiale Zhou}\textsuperscript{5}
\quad \textbf{Haohuai He}\textsuperscript{6}
\quad \textbf{Zhi Song}\textsuperscript{7}
\quad \textbf{Hanbo Huang}\textsuperscript{8}
\quad \textbf{Jiehui Huang}\textsuperscript{9}
\quad\textbf{Xun Lin}\textsuperscript{5}
\\[0.25em]
\textbf{Zhipeng Deng}\textsuperscript{5}
\quad\textbf{Zhaoxing Li}\textsuperscript{5}
\quad\textbf{Guanxing Chen}\textsuperscript{7}
\quad\textbf{Yaokun Li}\textsuperscript{10}
\quad\textbf{Mengran Li}\textsuperscript{10}
\\[0.25em]
\textbf{Songming Zhang}\textsuperscript{11}
\quad\textbf{Zhe Liu}\textsuperscript{1,*}
\\[0.55em]
\normalfont\textsuperscript{1}Zhejiang University
\quad \textsuperscript{2}The University of Hong Kong
\quad \textsuperscript{3}National University of Singapore
\\[0.25em]
\textsuperscript{4}Jiangnan University
\quad \textsuperscript{5}Westlake University
\quad \textsuperscript{6}The Hong Kong Polytechnic University
\\[0.25em]
\textsuperscript{7}City University of Hong Kong
\quad \textsuperscript{8}Shanghai Jiao Tong University
\\[0.25em]
\textsuperscript{9}The Hong Kong University of Science and Technology
\quad \textsuperscript{10}Sun Yat-sen University
\\[0.25em]
\textsuperscript{11}Nanjing University
\\[0.35em]
\normalfont\textsuperscript{\textdagger}Equal contribution
\quad
\normalfont\textsuperscript{*}Corresponding authors
\\[0.35em]
\normalfont Email: xiaogangxu00@gmail.com, zhe.liu@zju.edu.cn
}

\begin{document}

\noindent
\par\vspace{0.12in}
\maketitle

\begin{abstract}
Large language models can describe mechanisms, yet scalable post-training still depends on costly, manually curated biological reasoning traces. Here we show that cellular perturbation atlases can instead become reinforcement-learning environments, where measured gene responses provide computable rewards for biological reasoning. We introduce PertMind, which combines trusted-trajectory supervised initialization with gene-, pathway-, and format-level reinforcement signals. Although trained only on forward perturbation-response prediction, PertMind improves response inference in unseen cellular contexts while retaining general language capabilities. It also transfers, without task-specific post-training, to reverse perturbation identification, double-perturbation reasoning, phenotypic-screen prioritization, and biological-process interpretation. PertMind further generates biological profiles that support competitive gene, cell, and donor representations across multiscale downstream tasks. These results support the hypothesis that reinforcement on experimental endpoints can concentrate reusable biological strategies already accessible to pretrained models. More broadly, perturbation-derived reinforcement learning offers a scalable route for transforming expanding experimental atlases into training environments for general-purpose biological reasoning.
\par\medskip
\noindent\textbf{Project page:} \url{https://shapsider.github.io/PertMind/}
\end{abstract}

\section{Introduction}
\label{sec:introduction}

Predicting how a chemical perturbation reshapes a cell is central to mechanism-guided drug discovery, target prioritization, and functional genomics \citep{tang2023dsil,he2025dual,chen2026diconsite}. Useful predictions must link an intervention to pathway propagation, gene regulation, and cellular context across several biological scales \citep{tang2024modal,he2025sckan}. Large language models (LLMs) are attractive here because they internalize broad biomedical knowledge and can express mechanistic narratives in natural language. However, generic pretraining alone offers no direct signal on which biologically plausible narrative fits a concrete perturbation query \citep{yuan2026plausibility}. Perturbation-response systems addressed to this problem, including LLM-based reasoning harnesses \citep{wei2026vcworld} and non-LLM deep-learning baselines \citep{ahlmann2025deep}, train on well-defined endpoints or structured prediction tasks. They rarely optimize against the mechanism-oriented language that models actually generate. The bottleneck for reasoning-capable biology models is therefore supervision, not stored knowledge. In this work we report that public cellular perturbation atlases can be reorganized as scalable reinforcement-learning environments that turn measured interventions into reasoning experience for these models.

\begin{figure}[t]
\begin{center}
\includegraphics[width=\linewidth]{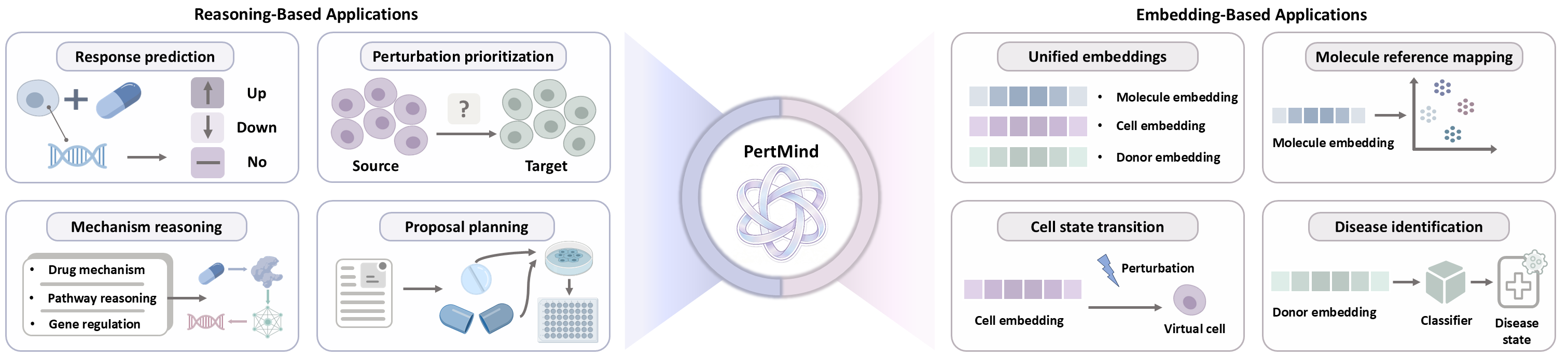}
\end{center}
\caption{Schematic scope of the capabilities enabled by PertMind. Reasoning-based applications use natural-language inference for perturbation-response prediction, perturbation prioritization, mechanism reasoning, and proposal planning. Embedding-based applications encode PertMind-generated biological profiles into reusable molecular, cellular, and donor representations for reference mapping, cellular perturbation-response prediction, and donor-level tumor-state reference mapping. The figure summarizes the intended capability space, not benchmark performance. Within this space, the quantitatively evaluated subset in the Results comprises forward perturbation-response prediction, reverse perturbation-condition inference, phenotypic-screen hit prioritization, biological-process naming, and molecular, cellular, and donor-level reference mapping.}
\label{fig:capability-overview}
\end{figure}

\begin{figure}[t]
\begin{center}
\includegraphics[width=\linewidth]{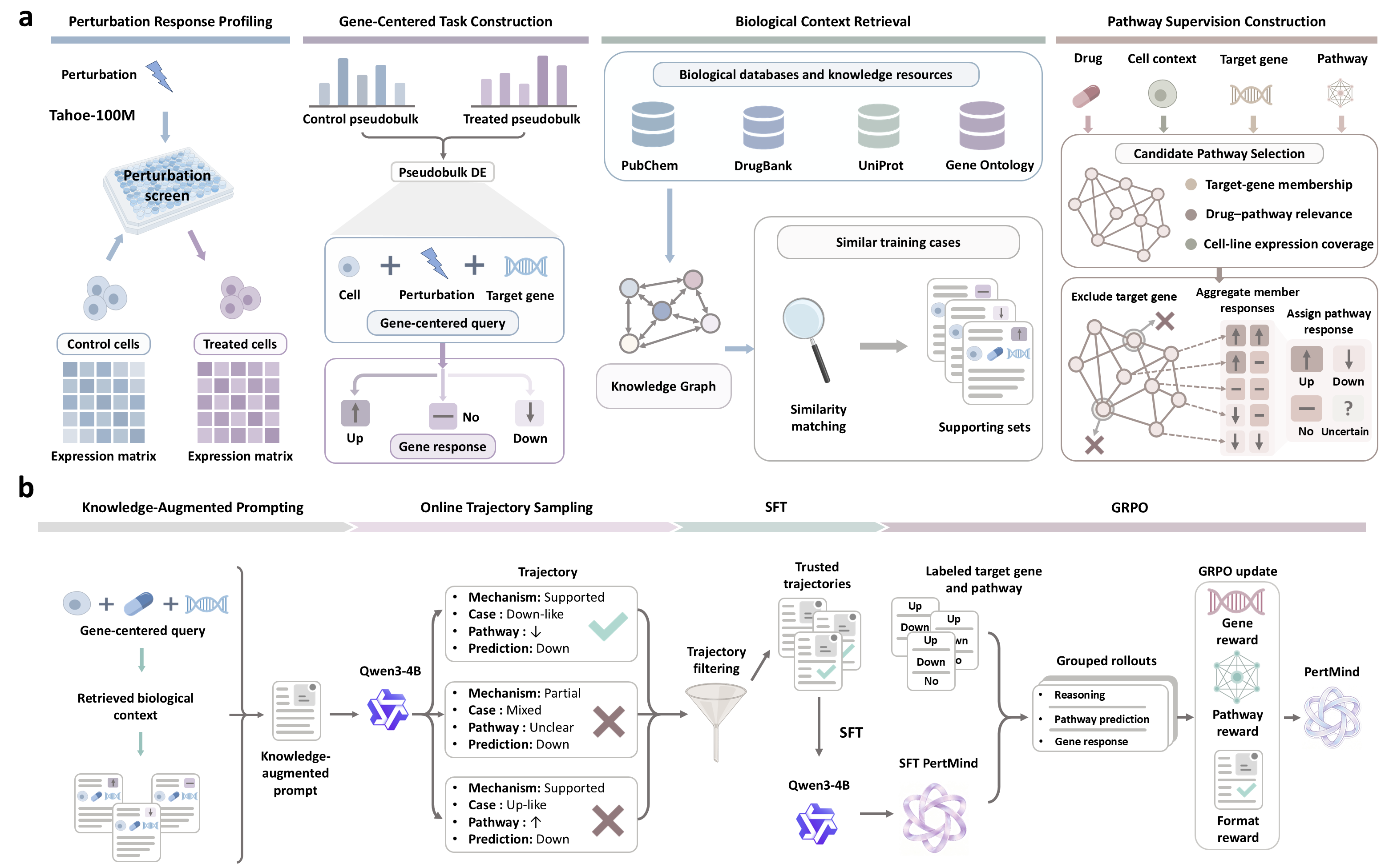}
\end{center}
\caption{Overview of PertMind. \textbf{(a)} Task and data construction. From each Tahoe-100M pseudobulk differential-expression record \citep{zhang2025tahoe} we assemble a gene-centered query \(x=(c,d,g)\) with an \(\mathrm{Up}\), \(\mathrm{Down}\), or \(\mathrm{No}\) label, retrieved biological context \(\mathcal{K}(x)\), and, when informative, outcome-independently selected candidate pathways with transcriptional pathway-response proxies. The \(\mathrm{uncertain}\) pathway outcome is a reward-side abstention state rather than a fourth model target, and receives no pathway reward. \textbf{(b)} Training. Knowledge-augmented prompts drive on-policy sampling from Qwen3-4B; trajectories that pass the trusted-trajectory filters seed one epoch of SFT; the resulting SFT policy is then optimized by pathway-supervised GRPO \citep{guo2025deepseek} under the composite gene, pathway, and format reward. The current query's target-gene label and all pathway proxy labels are reward-side information: they score sampled responses and are never placed in the model's prompt at training or inference time.}
\label{fig:overview}
\end{figure}

Recent biology-focused post-training narrows this gap by directly supervising model-generated reasoning trajectories. BioReason curates biological tasks that reward multimodal step-by-step inference \citep{fallahpour2025bioreason}, and Bio-KCoT augments long chain-of-thought generation with knowledge-graph facts for complex biomolecular problems \citep{lyu2025biokcot}. Both efforts show that biological reasoning can be trained, rather than left as a by-product of generic pretraining. Expert-authored trajectories and target rationales are, however, expensive to produce and difficult to scale across the drug--gene--context combinations that populate a perturbation atlas. They also fix in advance which intermediate steps count as good reasoning, so the supervision encodes assumptions about mechanism as much as about the correct answer. A complementary route is to leave the intermediate cascade unannotated and let public measurements evaluate whether a generated trajectory reaches the experimentally observed conclusion. The response remains hidden from the model as the quantity to predict, while its measured value is available only to the reward function after generation; no pre-specified stepwise labels are required.

Outcome-based reinforcement learning offers precisely this alternative to dense intermediate supervision. DeepSeek-R1 showed that policy optimization against terminal correctness can improve chain-of-thought reasoning without annotating every intermediate step \citep{guo2025deepseek}. Biology has begun to adopt the idea: OwkinZero applies verifiable-reward reinforcement learning to curated discovery-oriented question--answer tasks \citep{bigaud2025owkinzero}. Curated questions, however, still require a human author to decide which discovery-relevant items become training targets. Our proposal differs at the level of the reward interface itself. A perturbation experiment is an intervention with a directly measured terminal outcome, produced by the assay rather than by an annotator. A compound $d$ is applied to a cell line $c$, and a latent biological cascade unfolds. The response of the target gene $g$ is then scored as up-regulated, down-regulated, or not reliably changed. Every eligible triplet from a public atlas can be scored automatically, so the training environment scales with the atlas rather than with human curation.

This reward interface turns each query into a compressed observation of a latent biological cascade. Only the terminal endpoint is visible to the reward, yet that endpoint is shaped by drug mechanism, pathway propagation, gene regulation, and cellular context. Our central hypothesis is that reinforcement against these endpoints creates selection pressure for a policy that integrates all four factors rather than relying on drug identity, gene identity, or class base rates alone. Furthermore, Appendix~\ref{app:pre-rl-analysis} and Figure~\ref{fig:pre-rl-analysis} reveal clear optimization headroom: correct answers already occur in the Base model's sampling distribution, but greedy decoding and naive majority voting do not reliably select them. Endpoint-anchored reinforcement is therefore positioned to concentrate probability on the integrated strategies that yield correct conclusions.

PertMind operationalizes these two ideas starting from Qwen3-4B Base: perturbation endpoints provide scalable terminal supervision, and pathway-response checkpoints guide the policy toward an integrated intervention--pathway--gene--context representation. We assemble gene-centered queries from the Tahoe-100M perturbation atlas of roughly 100 million single-cell profiles \citep{zhang2025tahoe}. The atlas's precomputed differential-expression statistics provide terminal labels. A small corpus of trusted, model-generated trajectories provides a cold start through one epoch of supervised fine-tuning, establishing structured, evidence-grounded reasoning before reinforcement learning. Reinforcement learning then optimizes three complementary signals. A gene-level reward scores the experimentally observed $\mathrm{Up}$, $\mathrm{Down}$, or $\mathrm{No}$ endpoint. A pathway-level reward scores structured pathway-direction predictions derived from transcriptional response summaries. A format reward keeps outputs parseable. Every query-specific outcome remains hidden from the model.

Figure~\ref{fig:capability-overview} organizes the transferred capabilities into reasoning-based and embedding-based applications. For reasoning-based applications, PertMind approaches a task-specialized model on reverse perturbation-condition inference, extends to double-perturbation queries without double-perturbation training, improves phenotypic-screen hit prioritization through screen-specific briefs, and increases agreement with curated biological-process names. For embedding-based applications, its generated gene descriptions are encoded once, adapted per task, and composed into gene, cell, and donor representations that remain competitive with scale-specific representation models. We call a PertMind capability \emph{emergent}, in an operational sense, when it appears on a target task that was absent from post-training. Cross-task evidence under this definition provides the most informative test of whether PertMind acquires reusable biological strategies beyond its forward target-gene objective. In this study, our cross-task results support the acquisition of reusable biological reasoning strategies through reinforcement on perturbation endpoints.

\section{Methods}
\label{sec:methods}

We define PertMind's training interface as the condition-specific direction of a target gene's response, a computable projection of a multi-step biological process involving perturbation targeting, pathway propagation, and cell-context modulation. Given a cell line, a small-molecule perturbation, and a target gene, the policy predicts this discrete projection and states a mechanistic rationale that can be audited against retrieved biological evidence. The pipeline in Figure~\ref{fig:overview} starts from Qwen3-4B Base and proceeds in three stages: (i) construction of a perturbation-derived supervision corpus from precomputed Tahoe-100M statistics under strict cell-line-disjoint controls (Section~\ref{sec:dataset}); (ii) knowledge-augmented, on-policy trajectory sampling followed by one epoch of supervised fine-tuning (SFT), which produces the reinforcement-learning initialization \(\pi_{\theta_{\mathrm{SFT}}}\) (Section~\ref{sec:trajectory}); and (iii) pathway-supervised Group Relative Policy Optimization (GRPO) \citep{guo2025deepseek}, the principal optimization stage, in which a gene-level outcome reward is complemented by a verifiable, structured intermediate reward derived from transcriptional pathway-response proxies (Section~\ref{sec:grpo}). Reproducibility details deferred from this section are collected in Appendix~\ref{app:methods-details}.

\subsection{Problem formulation}
\label{sec:problem-formulation}

We represent a query as an ordered triplet of categorical identifiers
\begin{equation}
x = (c, d, g),
\end{equation}
where \(c\) denotes a cell-line identifier, \(d\) a small-molecule identifier, and \(g\) a target-gene identifier. These symbols index discrete biological entities rather than continuous scalar measurements or vector-valued features. The task label space is
\begin{equation}
\mathcal{Y} = \{\mathrm{Up}, \mathrm{Down}, \mathrm{No}\},
\end{equation}
indicating that the target gene is significantly up-regulated, significantly down-regulated, or shows no reliable evidence of differential expression under \(x\). This ternary label is a computable training interface rather than a definition of the model's capabilities: it captures the sign of the response that a multi-step regulatory process produces at the target gene, and we use it because it can be estimated from experimental data at scale and admits a well-defined reward. Given \(x\) and a biological context \(\mathcal{K}(x)\), assembled from external biological knowledge and training-split supporting cases, the policy \(\pi_\theta\) samples a reasoning trajectory \(z\) together with a structured final prediction \(\hat y_g\):
\begin{equation}
(z, \hat y_g) \sim \pi_\theta(\cdot \mid x, \mathcal{K}(x)).
\end{equation}
The natural-language trajectory \(z\) is an auditable explanation constructed from available knowledge and retrieved evidence. The experimentally derived label \(y_g\) anchors the correctness of the final gene-level conclusion, and pathway supervision (Section~\ref{sec:grpo}) provides additional, verifiable intermediate signal on top of \(y_g\).

\subsection{Perturbation-derived supervision}
\label{sec:dataset}

Training queries and labels are derived from the \texttt{pseudobulk\_differential\_expression} release in Tahoe-100M \citep{zhang2025tahoe}, which reports treated-control pseudobulk differential expression for each cell line, drug, concentration, and plate. PertMind consumes these precomputed DESeq2-style statistics directly and does not re-estimate primary labels from single-cell expression matrices. For each surviving triplet we write \(\Delta_{c,d,g}\) for \texttt{log2FoldChange}, \(q_{c,d,g}\) for \texttt{padj}, and \(\mu_{c,d,g}\) for \texttt{baseMean}, where dose- and plate-level records sharing a drug identity are collapsed to a single per-drug record by the rule detailed in Appendix~\ref{app:data-details}; records with missing \(\Delta\) or \(q\) and underpowered comparisons are discarded there.

Discrete labels are assigned by
\begin{equation}
y_g =
\begin{cases}
\mathrm{Up}, & q_{c,d,g} \leq 0.05 \ \land \ \Delta_{c,d,g} \geq 0.585, \\
\mathrm{Down}, & q_{c,d,g} \leq 0.05 \ \land \ \Delta_{c,d,g} \leq -0.585, \\
\mathrm{No}, & |\Delta_{c,d,g}| < 0.20 \ \land \ q_{c,d,g} \geq 0.50 \ \land \ \mu_{c,d,g} \geq 10, \\
\mathrm{ambiguous}, & \text{otherwise},
\end{cases}
\end{equation}
where \(0.585 \approx \log_2(1.5)\). Records labeled \(\mathrm{ambiguous}\) are excluded from training and evaluation, and \(\mathrm{No}\) denotes strong evidence of no differential expression at the stated thresholds rather than a claim of exact biological null effect. Each retained triplet additionally carries a confidence weight \(w^{\mathrm{conf}}_{c,d,g} \in [0.1, 1]\) computed from \((\Delta_{c,d,g}, q_{c,d,g}, \mu_{c,d,g})\); the closed forms and their derivation are given in Appendix~\ref{app:data-details}. The training corpus is
\begin{equation}
\mathcal{D}_{\mathrm{train}} = \{(c, d, g, y_g, w^{\mathrm{conf}}_{c,d,g}) : c \in \mathcal{C}_{\mathrm{train}}, \ y_g \in \mathcal{Y}\}.
\end{equation}

To match the VCWorld/GeneTAK target-domain protocol \citep{wei2026vcworld}, we assign every Tahoe-100M sample by cell line before constructing triplets and partition cell lines rather than triplets, \(\mathcal{C}_{\mathrm{train}} \cap \mathcal{C}_{\mathrm{test}} = \emptyset\). All samples from the five protocol-defined held-out lines C32, HepG2C3A, HOP62, Hs~766T, and PANC-1 form the evaluation domain; all samples from every other Tahoe-100M cell line form the development domain used for training and validation. These are disjoint cell-context domains within the same atlas, not separate datasets; drugs and genes may recur across domains, but cell lines cannot. The common eligibility and label rules above are then applied identically within each domain; no cell line, drug, or eligible triplet is subsampled. All thresholds, pathway-label parameters, retrieval rules, and model-selection decisions are fixed using only the development domain; the five test cell lines remain closed until final evaluation. Leakage controls are strict: the current query's target-gene label, its differential-expression statistics, and any pathway proxy label never enter the prompt; all retrieval pools are restricted to the training partition, always exclude the query and its repeated measurements, and never access validation or held-out-cell outcomes; pathway candidate selection (Section~\ref{sec:grpo}) uses only outcome-independent evidence; and no query-specific expression measurement is ever placed on the model side. By contrast, retrieved training cases may expose their own observed outcomes as historical evidence, including same-\((d,g)\) cases from another training cell line as cross-context evidence, but never the outcome of the current query, a validation query, or any held-out cell line. The grouped development split and the treatment of aggregated single-cell replicates are described in Appendix~\ref{app:data-details}.

\subsection{Knowledge-grounded RL initialization}
\label{sec:trajectory}

Reinforcement learning from a sparse three-class terminal reward alone provides little useful gradient when the initial policy rarely produces structured, evidence-grounded reasoning. We therefore first construct a small corpus of trusted, model-generated trajectories and use them for one epoch of SFT, giving GRPO a stable starting point.

For each training query, a retrieval module assembles the biological context \(\mathcal{K}(x)\) from an open-world biological knowledge graph over PubChem \citep{kim2023pubchem}, DrugBank \citep{knox2024drugbank}, UniProt \citep{uniprot2025}, Gene Ontology \citep{geneontology2023}, Reactome \citep{ragueneau2026reactome}, STRING \citep{szklarczyk2023string}, and CORUM \citep{tsitsiridis2023corum}, together with outcome-stratified supporting cases
\begin{equation}
\mathcal{S}(x) = \mathcal{S}_{\mathrm{Up}}(x) \cup \mathcal{S}_{\mathrm{Down}}(x) \cup \mathcal{S}_{\mathrm{No}}(x)
\end{equation}
drawn from \(\mathcal{C}_{\mathrm{train}}\); outcome-stratified retrieval prevents majority-class evidence from dominating the context without revealing the query's own label. The current Qwen3-4B policy then samples \(M\) candidate trajectories under \(\mathcal{K}(x)\). A candidate is retained only if it jointly satisfies: (i) its parsed final label matches \(y_g\); (ii) its output follows the required structure with a single unambiguous final answer; (iii) it does not claim access to hidden expression values or labels; (iv) invoked entities and relations are supported by the given context or explicitly flagged as inference; and (v) it contains a complete reasoning chain from perturbation target through pathway to gene-level conclusion. Trajectories passing all five criteria form the trusted-trajectory corpus \(\mathcal{D}_{\mathrm{traj}} = \{(x, z^\star, y_g)\}\). A query with no passing trajectory is excluded from \(\mathcal{D}_{\mathrm{traj}}\) and contributes no SFT loss, but it remains in \(\mathcal{D}_{\mathrm{train}}\) for GRPO, where experiment-derived rewards do not depend on trajectory acceptance.

We perform one epoch of SFT on \(\mathcal{D}_{\mathrm{traj}}\), reweighting the token-level cross-entropy by a Balanced Fine-Tuning (BFT) scheme \citep{tang2025aligning} that internally down-weights isolated low-confidence tokens while giving trajectories with at least one difficult local span larger relative sequence-level weights than easier trajectories. BFT changes how examples are weighted within standard SFT rather than adding a separate training stage; its objective and window definitions appear in Appendix~\ref{app:sft-details}. Because reinforcement learning is the principal optimization stage of PertMind, this stage is used only to establish structured output and mechanism-oriented reasoning behavior, and the resulting policy \(\pi_{\theta_{\mathrm{SFT}}}\) is retained both as the GRPO initialization and as the fixed reference for KL regularization.

\subsection{Pathway-supervised reinforcement learning}
\label{sec:grpo}

A terminal three-class reward collapses every response to final-label correctness and therefore cannot assign credit to intermediate biological consistency. PertMind complements this outcome reward with a structured intermediate reward derived from transcriptional pathway-response proxies, while keeping all model-visible information strictly outcome-independent. This combination of a gene-level outcome reward and a pathway-level checkpoint reward is the core method of PertMind.

For a subset of training triplets we build, before any \((c,d)\)-specific expression outcome is examined, a candidate pathway set \(\mathcal{P}^{\mathrm{cand}}(d,g)\) of at most three Reactome pathways containing \(g\), using the pathway annotation and mapping pipeline adopted in this work \citep{tang2025scprototransformer}, filtered by pathway size and ranked by outcome-independent target-gene and drug-mechanism evidence; the ranking terms are given in Appendix~\ref{app:pathway-details}. For each candidate we assign, at training time only, a transcriptional pathway-response proxy
\begin{equation}
y_P(c, d, g) \in \{\mathrm{Up}, \mathrm{Down}, \mathrm{No}, \mathrm{uncertain}\}
\end{equation}
that summarizes the coordinated transcriptional response of the non-target members \(P^{-g} = P \setminus \{g\}\). The proxy is computed within each full experimental condition \(\tilde d = (d, \mathrm{dose}, \mathrm{plate})\), where \(\mathrm{dose}\) is the administered concentration and \(\mathrm{plate}\) identifies the assay batch. All members contributing to a single pathway score therefore share the same dose and plate. The condition-level labels are then aggregated to \((c,d,g)\) by a conservative cross-condition consensus rule; the condition-level significance-weighted signed effect, aggregate score, sign consistency, active-member fraction, and the drug-level consensus are given in Appendix~\ref{app:pathway-details}. The target gene is excluded when forming \(P^{-g}\) because using its own expression to build the auxiliary reward would leak the primary answer into the process signal. The label \(y_P(c,d,g)\) is a reward-side transcriptional pathway-response proxy; it is not a direct measurement of protein activity, metabolic flux, or causal pathway activation, and it is used only for reward computation, never as model input. The prompt includes the full candidate set together with each pathway's name, its documented relation to the drug and target gene, and a small number of representative non-target members selected once per pathway using a pre-registered static score that combines annotation confidence and outcome-independent network centrality. It never includes any pathway's proxy direction, any member's expression statistics, or which candidates received a determinate \(y_P(c,d,g)\).

For each prompt \(x\) we sample a group of \(G\) responses from the current policy,
\begin{equation}
o_i \sim \pi_\theta(\cdot \mid x, \mathcal{K}(x)), \qquad i = 1, \dots, G,
\end{equation}
each containing a structured \texttt{pathway\_directions} field, a \texttt{mechanistic\_explanation} field, and a final gene-level label \(\hat y_{g,i}\). The terminal, primary-task reward is an unweighted binary indicator,
\begin{equation}
R_{\mathrm{gene}}(o_i) = \mathbf{1}[\hat y_{g,i} = y_g] \in \{0, 1\},
\end{equation}
which takes the value 0 for responses with no parseable label or contradictory labels. Let \(\mathcal{P}^{\mathrm{det}}(c,d,g) = \{P \in \mathcal{P}^{\mathrm{cand}}(d,g) : y_P(c,d,g) \neq \mathrm{uncertain}\}\) and \(K = |\mathcal{P}^{\mathrm{det}}(c,d,g)|\); the pathway reward averages the per-pathway indicator over the scored subset,
\begin{equation}
R_{\mathrm{pw}}(o_i) = \frac{1}{K} \sum_{P \in \mathcal{P}^{\mathrm{det}}(c,d,g)} \mathbf{1}[\hat y_{P,i} = y_P(c,d,g)] \in [0, 1],
\end{equation}
with \(R_{\mathrm{pw}} := 0\) when \(K = 0\) and a pathway mask \(m_{\mathrm{pw}} = \mathbf{1}[K > 0]\). The response is required to declare a direction for every candidate in \(\mathcal{P}^{\mathrm{cand}}(d,g)\); \(\mathrm{uncertain}\) candidates enter neither the sum nor the denominator, and, because the model is not told which candidates are scored, it cannot allocate effort selectively. Thus, \(R_{\mathrm{pw}}\) evaluates only the structured pathway slots and does not validate the free-text mechanistic explanation. The total reward is
\begin{equation}
R(o_i) = R_{\mathrm{gene}}(o_i) + \lambda_{\mathrm{pw}} \, m_{\mathrm{pw}} \, R_{\mathrm{pw}}(o_i) + \lambda_{\mathrm{fmt}} \, R_{\mathrm{fmt}}(o_i),
\end{equation}
The format reward is the conjunction of three explicit checks:
\begin{equation}
\begin{aligned}
I_{\mathrm{fields}}(o_i) &= \mathbf{1}[\text{all required fields are present}],\\
I_{\mathrm{vocab}}(o_i) &= \mathbf{1}[\text{all labels use the legal vocabulary}],\\
I_{\mathrm{final}}(o_i) &= \mathbf{1}[\text{exactly one terminal answer is parseable}],\\
R_{\mathrm{fmt}}(o_i) &= I_{\mathrm{fields}}(o_i) I_{\mathrm{vocab}}(o_i) I_{\mathrm{final}}(o_i) \in \{0,1\}.
\end{aligned}
\end{equation}
The auxiliary weights are chosen to satisfy \(\lambda_{\mathrm{pw}} > 0\), \(\lambda_{\mathrm{fmt}} > 0\), and \(\lambda_{\mathrm{pw}} + \lambda_{\mathrm{fmt}} < 1\). This inequality is decidable from the chosen weights alone and holds for every response, so no combination of auxiliary rewards can outrank a correct gene label: a response satisfying all auxiliary criteria but predicting the wrong gene label always scores strictly below one predicting the correct label. The component \(R_{\mathrm{pw}}\) directly supervises discrete pathway-direction predictions in the structured \texttt{pathway\_directions} field; it does not verify every free-text mechanistic claim, which remains outside the reward boundary and distinct from the trusted-trajectory validators used before SFT. Moreover, \(R_{\mathrm{pw}}\) does not require the target-gene response to match the pathway aggregate direction, because negative feedback, bypass compensation, and post-transcriptional regulation can decouple a target gene from its pathway's trend. The primary component \(R_{\mathrm{gene}}\) is therefore evaluated independently against the experiment-derived target-gene label \(y_g\) defined in Section~\ref{sec:dataset}.

\paragraph{Group-relative optimization.}
GRPO forgoes a learned value network and standardizes the total reward within each group,
\begin{equation}
\hat A_i = \frac{R(o_i) - \operatorname{mean}_j R(o_j)}{\operatorname{std}_j R(o_j) + \epsilon},
\end{equation}
where \(\epsilon>0\) is a numerical stabilizer. Updates therefore depend on the relative quality of responses to the same query and are less sensitive to reward-scale differences across drugs or cell lines. A group in which every response has the same total reward has zero variance and contributes no gradient. If only \(R_{\mathrm{gene}}\) is uniform, the auxiliary terms can still rank responses and provide partial credit without reversing any preference established by gene-label correctness.

\paragraph{Confidence weighting and policy regularization.}
After group standardization, the per-triplet confidence \(w^{\mathrm{conf}}_{c,d,g}\) multiplies the complete sample objective rather than \(\hat A_i\) itself. The multiplier is shared by all responses in a group, so it scales the triplet's parameter update without changing the ordering induced by \(\{\hat A_i\}_{i=1}^{G}\). PertMind then combines this unscaled standardized advantage with a clipped importance-weighted surrogate against the behavior policy \(\pi_{\theta_{\mathrm{old}}}\) and a sampled-token KL penalty toward the frozen SFT reference \(\pi_{\theta_{\mathrm{SFT}}}\). The full per-response objective, sampled-token KL estimator, and reward-hacking monitors are given in Appendix~\ref{app:grpo-details}. Regularizing every RL update toward the frozen SFT policy helps preserve the linguistic competence and reasoning structure established during the cold start. We refer to the model obtained after this confidence-weighted, KL-regularized GRPO stage as PertMind.

\section{Datasets and Experimental Setup}
\label{sec:experimental-setup}

\subsection{Training data and leakage-controlled split}

We use the complete Tahoe-100M perturbation atlas \citep{zhang2025tahoe} as the source of PertMind's forward perturbation supervision and adopt the five-cell-line outer split defined by the VCWorld/GeneTAK evaluation protocol \citep{wei2026vcworld}. Before any filtering or label construction, every sample is assigned by cell-line identity. All samples belonging to C32, HepG2C3A, HOP62, Hs~766T, or PANC-1 are reserved for final evaluation; all samples from the remaining Tahoe-100M cell lines are assigned to the development domain. Thus, development and evaluation data come from the same atlas but are disjoint in cell context. Within the development domain, retained \((c,d,g)\) groups are split 90:10 into training and validation partitions, stratified by cell line and three-class label, with all records sharing a \((c,d,g)\) key kept in the same partition. The validation partition is used only for model selection, threshold calibration, and early stopping. The same condition-level quality controls and the same \(\mathrm{Up}/\mathrm{Down}/\mathrm{No}\) definitions are applied to development and test domains, and ambiguous records are excluded as specified in Section~\ref{sec:dataset}. We do not downsample cell lines, drugs, or eligible triplets. Retrieval uses training records only, and neither validation outcomes nor any outcome from the five held-out cell lines is available to trajectory generation, SFT, GRPO, pathway-candidate selection, or hyperparameter selection. Drugs and genes may appear in both domains, but every evaluation cell context is unseen during training.

\subsection{Training configuration and model variants}

PertMind is initialized from Qwen3-4B Base. For trusted-trajectory construction, we draw \(M=8\) candidate responses per training query and retain only trajectories satisfying all validators in Section~\ref{sec:trajectory}. BFT-weighted SFT is run for one epoch with LoRA rank 64, LoRA alpha 128, a learning rate of \(1\times10^{-4}\), micro-batch size 1, gradient accumulation over 16 steps, and a maximum sequence length of 10,240 tokens. GRPO is then run for three epochs with \(G=8\) responses per prompt, LoRA rank 64, LoRA alpha 128, learning rate \(1\times10^{-5}\), per-device batch size 2, gradient accumulation over 4 steps, maximum completion length 2,048, temperature 0.9, and top-\(p=0.95\). We use \(\lambda_{\mathrm{pw}}=0.15\), \(\lambda_{\mathrm{fmt}}=0.05\), KL coefficient \(\beta=0.01\), and clipping range \(\varepsilon=0.20\). Pathway candidates use \(\alpha_{\mathrm{gene}}=0.60\) and \(\alpha_{\mathrm{drug}}=0.40\); pathway proxies use \(q_{\min}=10^{-6}\), \(a_{\max}=6\), \(\tau_{\mathrm{dir}}=0.20\), \(\tau_{\mathrm{no}}=0.05\), \(\gamma=0.60\), and \(\eta_{\mathrm{no}}=0.20\). All settings are selected on the training/validation domain and frozen before final evaluation.

The ablation study compares six Qwen3-4B-derived variants under the same held-out-cell protocol: the unmodified Base model; BFT-weighted SFT; a shuffled-pathway sanity control; gene-only GRPO with the terminal outcome reward; prompt-only pathway context without pathway reward; and full PertMind with gene, pathway, and format rewards. The shuffled control preserves pathway frequency and prompt structure while permuting pathway labels within the training partition. Appendix~\ref{app:ablation-variants} specifies the initialization, model-visible pathway context, and active reward terms for every variant.

\subsection{Evaluation datasets, tasks, and baselines}

\paragraph{Forward perturbation response and general capability.}
We use the open-source VCWorld/GeneTAK protocol \citep{wei2026vcworld} on all eligible triplets from the five held-out Tahoe-100M cell lines. Differential-expression detection (DE) predicts whether a target gene changes, and direction prediction (DIR) distinguishes up- from downregulation among changed genes. We report Accuracy and Macro-F1 and compare Random, graph attention networks (GAT) \citep{velickovic2018graph}, CPA \citep{lotfollahi2023predicting}, scVI \citep{lopez2018generative}, STATE \citep{adduri2025predicting}, the VCWorld harness instantiated with Gemini-2.5-Flash, and PertMind. General-capability retention is evaluated on the standard 57-subject MMLU benchmark \citep{hendrycks2021mmlu} and 67-subject CMMLU benchmark \citep{li2024cmmlu} under zero-shot and five-shot prompting, comparing Base, SFT, and PertMind by multiple-choice accuracy.

\paragraph{Perturbation-condition prediction.}
The reverse task receives source and target cell states, converts their transition into up- and downregulated DEGs, and ranks candidate interventions. On the primary-human-T-cell CRISPRa/CRISPRi dataset of Schmidt et al.\ \citep{schmidt2022crispr}, each query ranks 69 candidate perturbations; we report Top-1 accuracy, Top-5 accuracy, and F1 and compare CellNavi \citep{wang2025cellnavi}, PertMind, SFT, Base, DEG ranking, GEARS \citep{roohani2023gears}, and Random. The combinatorial evaluation uses 131 double-perturbation pairs from the Norman Perturb-seq dataset \citep{norman2019exploring} and a pool of 105 single-gene candidates. For each pair, we retain the separate ranks of both ground-truth genes and compare PertMind with CellNavi and GEARS; no double-perturbation example is used to adapt PertMind for this task.

\paragraph{Cross-task biological reasoning.}
AssayBench \citep{de2026assaybench} evaluates gene prioritization from free-text phenotypic-screen protocols. We use its official year-fold0 test split of 334 screens and adjusted NDCG at 100 (AnDCG@100) as the primary metric; Hits@100 and hallucination rate are auxiliary measures in the 36-screen stratified BKI ablation. Protocol-only ranking is compared with format-only, generic-note, mismatched-brief, and matched PertMind-brief conditions. The full test evaluates protocol-only and PertMind-augmented GPT-5.4, Gemini 3 Pro, and Qwen3.5-397B-A17B across five phenotype classes. For biological-process naming, we follow GeneAgent \citep{wang2025geneagent} on 1,000 Gene Ontology, 50 NeST, and 56 MSigDB gene sets. GPT-4 and GeneAgent are compared with their PertMind-augmented counterparts using ROUGE-L/1/2 and MedCPT \citep{jin2023medcpt} semantic-similarity percentile rank against 12,320 background process terms.

\paragraph{Multi-scale representation.}
At the molecular level, GenePT \citep{chen2025simple}, scGPT \citep{cui2024scgpt}, scProtoTransformer \citep{tang2025scprototransformer}, and PertMind gene embeddings are evaluated by logistic regression and random forests on four gene-property reference-mapping tasks, using AUC. At the cellular level, scProtoTransformer, STATE-SE, STATE-SM, and PertMind representations are connected to the same STATE-based expression decoder \citep{adduri2025predicting} and evaluated on HepG2, K562, and RPE1 perturbation data using DES, PDS, and MAE. At the donor level, the pan-cancer reference-mapping setup of scTransMIL \citep{tang2025sctransmil} compares DEGAS \citep{johnson2022degas}, KIDA \citep{tang2024kida}, scIDST \citep{wehbe2025scidst}, scVI with attention-based multiple-instance learning \citep{lopez2018generative,ilse2018attention}, scTransMIL variants, and PertMind at 50\%, 75\%, and 100\% of the reference-training donors, using Accuracy, AUC, and F1. Each task-specific projection is reinitialized within each training fold and jointly optimized with the downstream classifier or decoder using only that fold's reference-training partition.

\subsection{Metrics and statistical reporting}

Unless a benchmark defines a different official protocol, Accuracy and F1 use their standard definitions, Macro-F1 averages class-wise F1 without prevalence weighting, and ranking metrics are computed over the complete candidate list. Higher values are better for Accuracy, F1, AUC, DES, PDS, ROUGE, and AnDCG; lower values are better for MAE and perturbation rank. Error bars report standard deviation over five independent runs for forward response, general capability, Schmidt reverse prediction, and cell-level representation experiments, and over seven runs for donor-level mapping. GeneAgent ROUGE error bars follow its nine batch-sampling replicates. These error bars are descriptive measures of run-to-run variation; captions explicitly identify panels without replicated-run uncertainty.

\section{Results}
\label{sec:results}

\subsection{PertMind improves perturbation-response prediction}
\label{sec:results-perturbation-prediction}

\begin{figure}[t]
\begin{center}
\includegraphics[width=\linewidth]{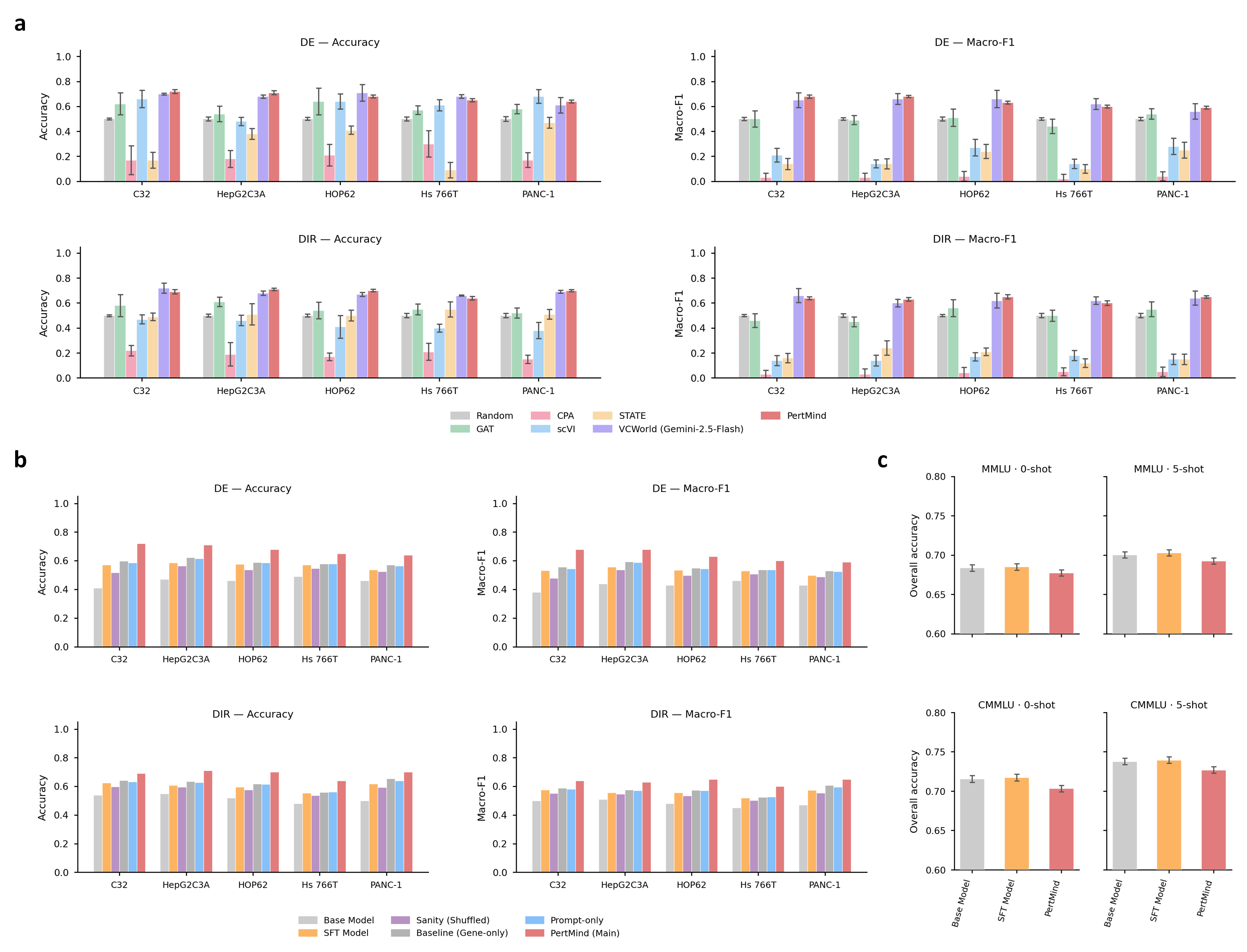}
\end{center}
\caption{Perturbation-response prediction, component ablation, and general-capability evaluation. \textbf{(a)} VCWorld benchmark accuracy and Macro-F1 for differential-expression detection (DE) and direction prediction (DIR) on the five unseen cell lines, comparing Random, GAT, CPA, scVI, STATE, VCWorld (Gemini-2.5-Flash), and PertMind. \textbf{(b)} Ablation across Base, SFT, shuffled-pathway sanity, gene-only outcome RL, prompt-only pathway context, and the complete PertMind pipeline, shown per cell line for each of the DE and DIR Accuracy and Macro-F1 combinations; this panel does not display run-level error bars. \textbf{(c)} Overall MMLU and CMMLU accuracy under 0-shot and 5-shot evaluation for Base, SFT, and PertMind. Error bars in panels (a) and (c) denote standard deviation over five independent runs.}
\label{fig:results-overview}
\end{figure}

We first evaluated PertMind on the open-source VCWorld perturbation-response benchmark, whose evaluation protocol defines the DE and DIR tasks on unseen cell lines (Figure~\ref{fig:results-overview}a). Within this framework, we compared PertMind against the VCWorld reference harness configured with the closed-source Gemini-2.5-Flash model. PertMind matched or modestly exceeded this Gemini-based reference on average across both tasks and evaluation metrics, and it outperformed the reference in a majority of the held-out cell-line settings. A Qwen3-4B-derived PertMind therefore reached the performance range of a substantially larger closed-source model and exceeded it in individual unseen contexts. These results support the hypothesis that reinforcement learning on cellular perturbation supervision improves biological perturbation reasoning, without by themselves establishing an emergent capability or a biological causal mechanism.

To attribute this gain to a specific component, we compared six training variants across all task--metric combinations and held-out cell lines (Figure~\ref{fig:results-overview}b). SFT produced the first substantial improvement over the Base model, and gene-only outcome RL provided a further, albeit modest, gain, establishing a contribution from gene-level supervision. Adding pathway context to the prompt without a corresponding pathway reward did not improve over gene-only RL. By contrast, combining the gene-level outcome reward with biologically aligned pathway supervision produced the best-performing PertMind variant, while shuffling the pathway labels reduced performance below the SFT model. These results indicate that gene-level and pathway-level rewards make complementary contributions: the former anchors the final perturbation outcome, whereas the latter supplies structured intermediate supervision that cannot be replaced by pathway text alone.

We next asked whether pathway-supervised RL came at the cost of general language capability. Figure~\ref{fig:results-overview}c and Appendix Figure~\ref{fig:general-capability} report overall and per-subject accuracy on MMLU and CMMLU under zero-shot and few-shot evaluation. PertMind remained close to both the Base and SFT models across these settings, with only small and consistent reductions in overall accuracy. Two design choices may have contributed to this retention. First, every RL update included a KL penalty toward the frozen SFT policy, directly constraining policy drift throughout GRPO; this self-distillation-like reference constraint is consistent with evidence that self-distillation can support continual learning \citep{shenfeld2026self}. Second, BFT attenuated gradients from locally low-confidence tokens while balancing sequence-level contributions during the SFT cold start, reducing pressure to fit unreliable token targets before RL began. General capabilities were therefore largely retained after pathway-supervised RL, and we did not observe catastrophic forgetting.

Finally, we illustrated PertMind's behavior under a free-form natural-language query outside the strict training schema, asking about Vemurafenib treatment of C32 cells and its effect on MKI67. PertMind predicted a differential-expression event with a Down direction, and its response linked BRAF V600E inhibition to reduced MAPK/ERK signaling, lower proliferative activity, and reduced MKI67 expression. A literature audit of this response, presented in Appendix Figure~\ref{fig:natural-language-case}, associated its major mechanistic links with published sources. We present this case as an illustrative example of model behavior rather than as quantitative evidence of generalization or proof that the underlying reasoning is faithful.

\subsection{PertMind generalizes to perturbation-condition prediction}
\label{sec:results-perturbation-condition}

\begin{figure}[t]
\begin{center}
\includegraphics[width=\linewidth]{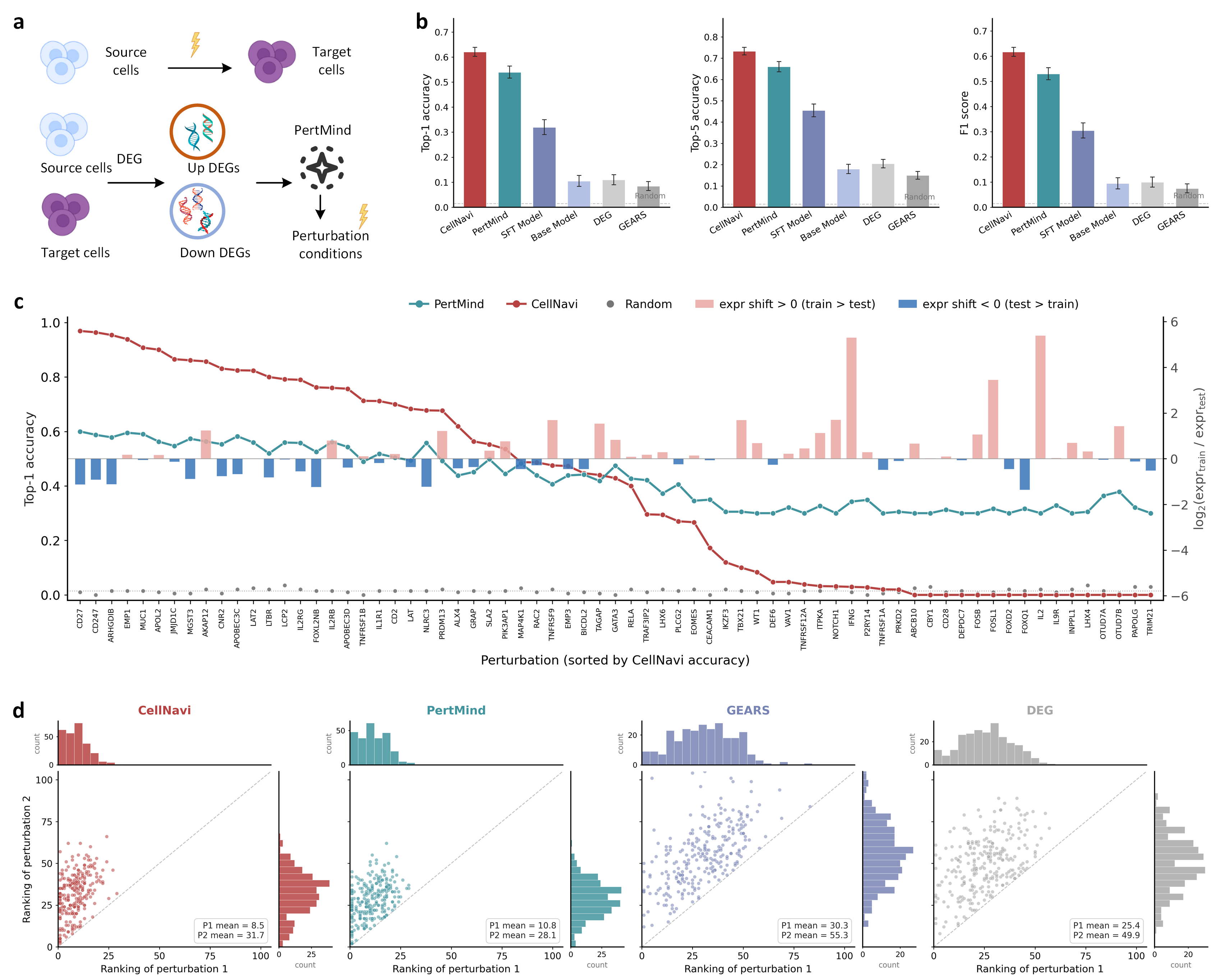}
\end{center}
\caption{Perturbation-condition prediction from cellular transitions. \textbf{(a)} Task schematic. In contrast to forward perturbation-response prediction, the reverse task provides source and target cells, summarizes their transition as up- and downregulated DEGs, and asks the model to rank candidate perturbation conditions. \textbf{(b)} Top-1 accuracy, Top-5 accuracy, and F1 score on the Schmidt primary T-cell dataset, comparing CellNavi, PertMind, SFT, Base, DEG, GEARS, and Random. Error bars denote standard deviation over five independent runs. \textbf{(c)} Per-perturbation Top-1 accuracy on the Schmidt dataset, with perturbations ordered by CellNavi accuracy. Bars show the expression shift \(\log_2(\mathrm{expression}_{\mathrm{train}}/\mathrm{expression}_{\mathrm{test}})\); positive values indicate higher expression in the training state. \textbf{(d)} Joint rankings of the two ground-truth perturbations in the Norman double-perturbation dataset. Lower ranks are better; marginal histograms show the rank distributions.}
\label{fig:perturbation-condition}
\end{figure}

The preceding experiments evaluated the forward problem: given a cellular context and perturbation, predict the resulting transcriptional response. We next inverted this problem and asked which perturbation condition could induce an observed transition from source to target cells (Figure~\ref{fig:perturbation-condition}a). We represented each transition by its up- and downregulated DEGs and prompted PertMind in natural language to rank candidate perturbations. This evaluation required no additional task-specific fine-tuning of PertMind. By comparison, CellNavi uses a driver-gene predictor fine-tuned on CRISPR-screen data for cellular-transition inference \citep{wang2025cellnavi}.

We first evaluated this reverse task on primary T-cell CRISPR perturbations from the Schmidt dataset \citep{schmidt2022crispr}. Although PertMind had been post-trained only for forward perturbation-response prediction, it approached the performance of the task-specialized CellNavi model across Top-1 accuracy, Top-5 accuracy, and F1 score (Figure~\ref{fig:perturbation-condition}b). It also substantially outperformed the Base and SFT models. Perturbation-condition prediction never appeared in the post-training corpus, and the gain over both controls emerged only after outcome RL. Under our operational definition, this pattern provides direct evidence of an emergent cross-task capability, while not by itself proving the internal mechanism responsible for the transfer.

Aggregate performance concealed a marked difference in perturbation-level robustness (Figure~\ref{fig:perturbation-condition}c). CellNavi performed strongly on many perturbations but varied sharply across conditions, with its accuracy collapsing toward zero for a subset of targets. These failures were concentrated among perturbations expressed more highly in the training state than in the test state. PertMind had a lower peak accuracy but maintained a substantially flatter performance profile across the ordered perturbations and across both directions of expression shift. This stability suggests broader generalization across unseen perturbation conditions and cell-state shifts, complementing CellNavi's stronger average performance.

We then tested whether this transfer extended to combinatorial interventions using the Norman Perturb-seq dataset \citep{norman2019exploring}. For each double perturbation, we examined the ranks assigned to both ground-truth conditions rather than crediting recovery of only the easier target (Figure~\ref{fig:perturbation-condition}d). CellNavi ranked the easier first perturbation more highly, whereas PertMind improved the rank of the second perturbation and reduced the imbalance between the two targets. One possible explanation is the training regime: multigene perturbations were excluded when CellNavi was trained for this evaluation \citep{wang2025cellnavi}, whereas PertMind could draw on the compositional reasoning prior of an LLM without double-perturbation-specific training. Together, these results show that post-training an LLM on forward cellular responses can transfer to reverse and combinatorial perturbation inference, suggesting a complementary prompt-based route for these settings.

\subsection{PertMind improves phenotypic-screen hit prioritization}
\label{sec:results-assaybench}

\begin{figure}[t]
\begin{center}
\includegraphics[width=\linewidth]{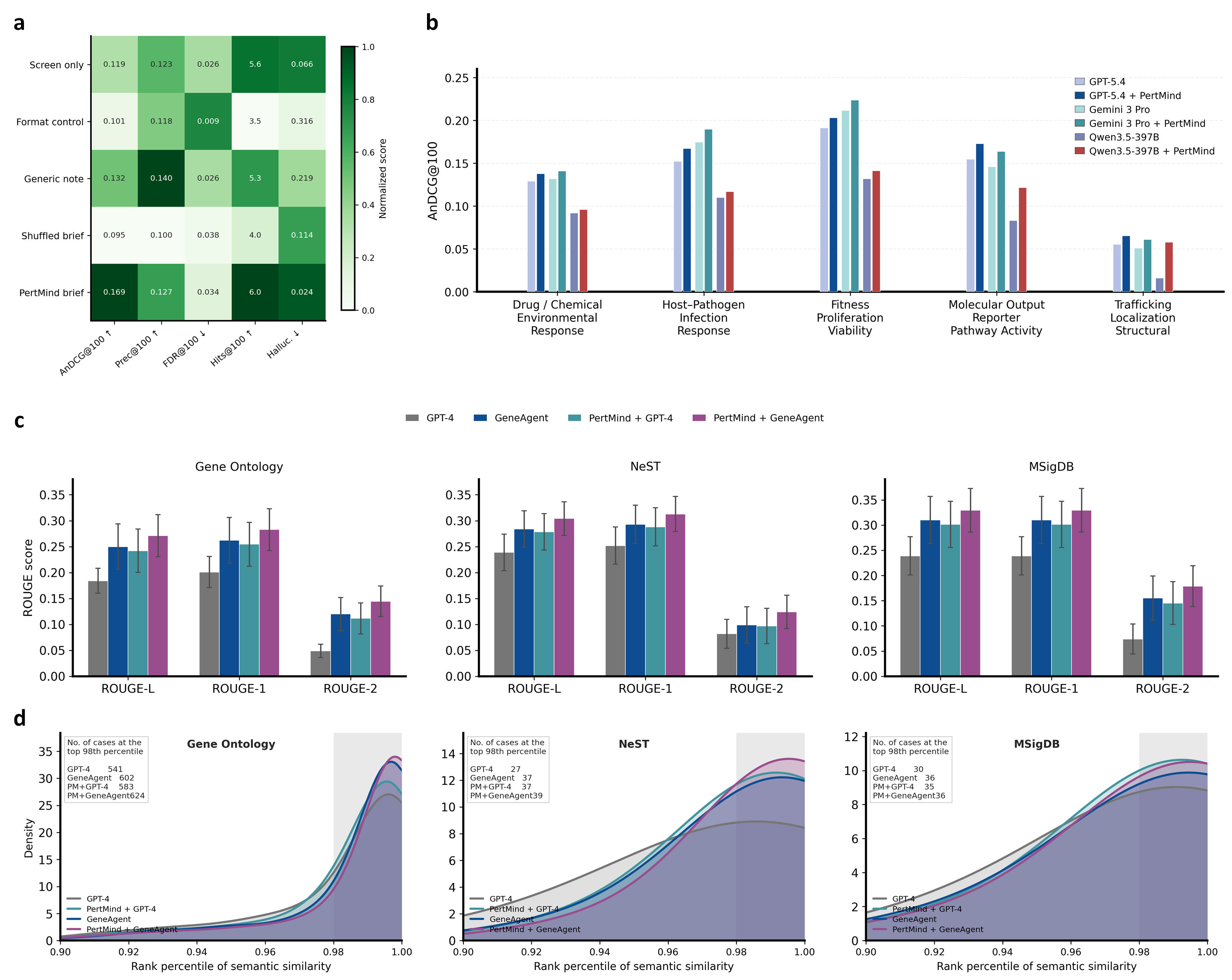}
\end{center}
\caption{Cross-task biological knowledge transfer with PertMind. \textbf{(a)} Biology Knowledge Injection (BKI) ablation on a stratified 36-screen subset of the AssayBench test set. Qwen3.5-397B-A17B ranks genes from the screen protocol alone or with a format control, generic note, mismatched PertMind brief, or matched PertMind brief. Cells report raw values; color denotes within-column normalization, while arrows indicate the preferred metric direction. \textbf{(b)} AnDCG@100 on the complete AssayBench year-fold0 test set (\(n=334\)), stratified by five coarse phenotype classes. GPT-5.4, Gemini 3 Pro, and Qwen3.5-397B-A17B (abbreviated as Qwen3.5-397B in the panel) are evaluated with the protocol alone and with a screen-specific PertMind brief. \textbf{(c)} ROUGE-L, ROUGE-1, and ROUGE-2 agreement between generated and ground-truth process names for 1,000 Gene Ontology, 50 NeST, and 56 MSigDB gene sets. Error bars denote s.d. estimated from nine batch-sampling replicates, using batch sizes of 200 for Gene Ontology and 20 for NeST and MSigDB. \textbf{(d)} Density of MedCPT semantic-similarity percentile ranks between generated and ground-truth process names against 12,320 annotated background terms. Only the top decile is shown; shading marks the \(\geq 98\)th-percentile region, and inset values give the number of gene sets in this region.}
\label{fig:cross-task-biology}
\end{figure}

We next asked whether the biological knowledge acquired from perturbation-response training could improve a distinct functional-genomics task. AssayBench casts each phenotypic CRISPR screen as gene prioritization from a free-text experimental protocol: a model returns a ranked list of 100 candidate genes, which is evaluated against experimentally defined hits using adjusted NDCG at 100 (AnDCG@100) \citep{de2026assaybench}. We used the official temporally split year-fold0 test set, which contains 334 screens spanning five coarse phenotype classes. To transfer PertMind's knowledge without updating the ranking model, we developed a two-stage Biology Knowledge Injection (BKI) procedure. PertMind first converts the protocol into a screen-specific biology brief containing mechanistic hypotheses, pathway cues, and directional constraints. A frontier ranking model then conditions on both the original protocol and this brief to produce the final gene ranking.

We isolated the source of the BKI gain on a stratified subset of 36 screens (Figure~\ref{fig:cross-task-biology}a). A matched PertMind brief achieved the strongest AnDCG@100, recovered the most hits within the top 100, and produced the lowest hallucination rate. Format-only text and a brief sampled from another screen did not reproduce this ranking gain, while a generic note provided a smaller improvement. The matched brief was not uniformly optimal for every auxiliary metric, but its advantage on the primary ranking metric, together with the mismatched-brief control, indicates that the benefit depended on screen-specific biological content rather than prompt length or formatting alone.

We then evaluated BKI on all 334 test screens and replaced the ranking backbone while keeping the knowledge-generation stage fixed (Figure~\ref{fig:cross-task-biology}b). Adding the PertMind brief improved AnDCG@100 for GPT-5.4, Gemini 3 Pro, and Qwen3.5-397B-A17B across all five phenotype classes. The consistent direction of improvement across proprietary and open-weight backbones shows that PertMind's contribution was not tied to a particular model family. It instead functioned as a transferable biological knowledge module for phenotypic-screen prioritization.

\subsection{PertMind enhances biological-process inference}
\label{sec:results-process-inference}

To test a second form of cross-task transfer, we adopted the GeneAgent benchmark for assigning biological process names to gene sets \citep{wang2025geneagent}. The benchmark comprises gene sets from Gene Ontology, NeST, and MSigDB, and compares each generated process name with a manually annotated reference. We augmented both GPT-4 and GeneAgent with biological context generated by PertMind. This setting tests whether PertMind can support process-level interpretation despite being trained with gene- and pathway-level rewards for perturbation responses.

PertMind improved lexical agreement with the reference process names across all three data sources (Figure~\ref{fig:cross-task-biology}c). PertMind-augmented GPT-4 approached the ROUGE performance of GeneAgent, whereas augmenting GeneAgent produced a further but smaller gain. This diminishing marginal improvement suggests that PertMind already supplied part of the biological knowledge otherwise obtained through GeneAgent's domain-database interactions. ROUGE measures token and phrase overlap, however, and therefore establishes agreement with curated labels rather than the faithfulness of the model's internal reasoning.

We next evaluated semantic rather than lexical agreement using MedCPT biomedical text embeddings \citep{jin2023medcpt}. For each gene set, we ranked the similarity between the generated and reference names against 12,320 annotated biological-process terms (Figure~\ref{fig:cross-task-biology}d). PertMind augmentation shifted predictions toward the highest background percentiles for both GPT-4 and GeneAgent, with dataset-specific variation in the magnitude of the gain. Thus, the improvement was not limited to reproducing the surface form of a gold label: PertMind also increased the frequency with which the reference process was among the closest semantic interpretations. These percentile ranks measure semantic alignment, not factual verification, and should be interpreted alongside the ROUGE results rather than as direct evidence of a faithful reasoning trace.

Across AssayBench and the GeneAgent benchmark, PertMind improved black-box frontier models through natural-language knowledge exchange without modifying their parameters. This use is conceptually related to the model-as-memory paradigm, in which a dedicated model exposes internalized domain knowledge to a separate executive model \citep{quek2026memo}. Related memory systems emphasize complementary aspects of maintaining useful state: resolving updates before evidence traversal \citep{song2026contract}, gating persistent and recent context at sequence boundaries \citep{huang2026unityshots}, and retaining verifier signals to govern subsequent agent memory \citep{wang2026memguard}. Our implementation does not reproduce these systems' memory-training, update-resolution, or governance protocols; instead, it shows that a perturbation-trained model can serve as a plug-and-play source of biological context. The resulting gains suggest that PertMind may reduce reliance on repeated external-database access for some tasks, although they do not establish complete replacement of curated resources. Together with perturbation-condition inference, these results are consistent with emergent biological transfer beyond PertMind's training objective.

\subsection{PertMind enables unified representations across molecular, cellular, and donor scales}
\label{sec:results-multiscale-representation}

\begin{figure}[t]
\begin{center}
\includegraphics[width=\linewidth]{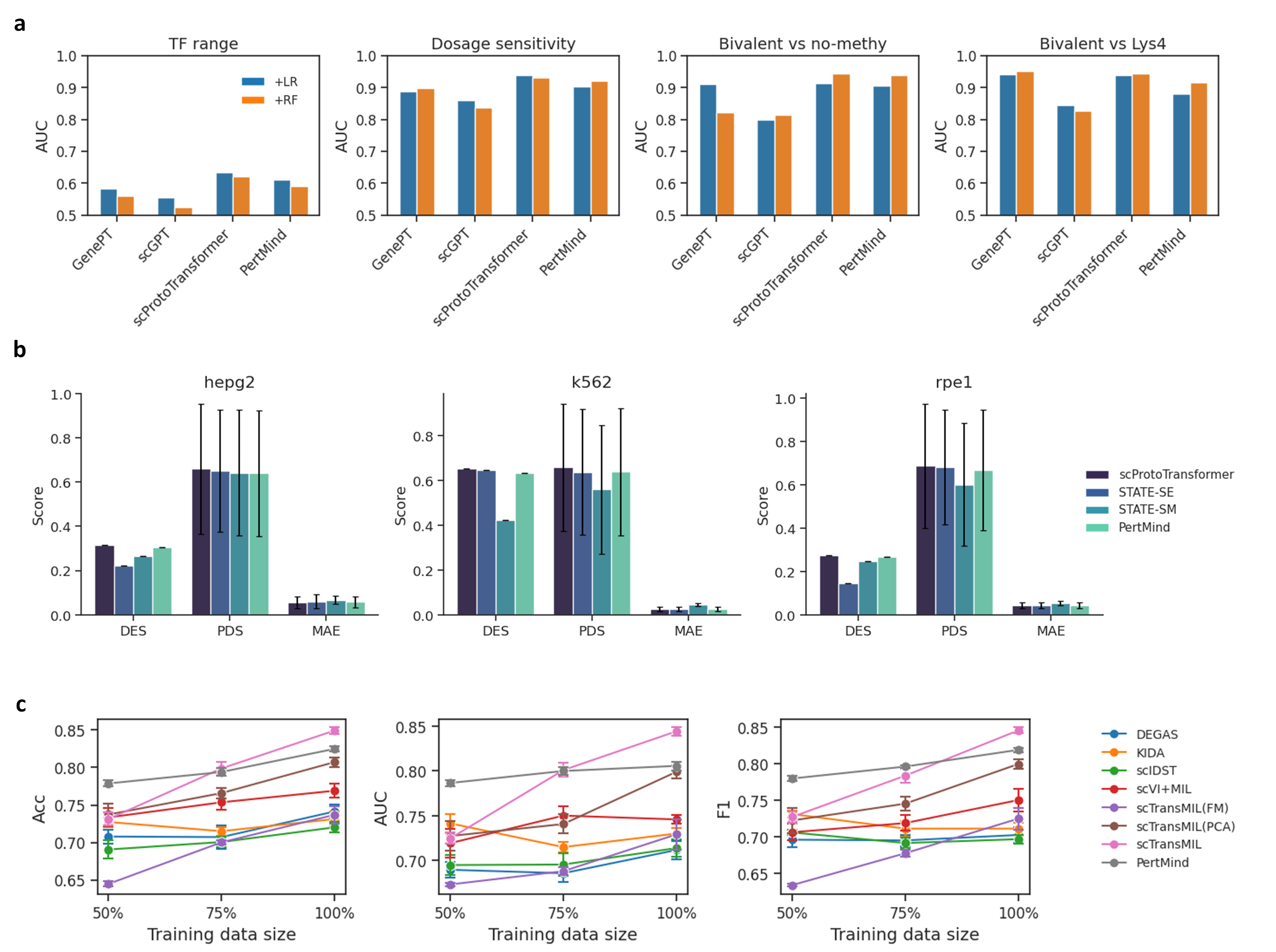}
\end{center}
\caption{Evaluation of PertMind-derived representations across biological scales. \textbf{(a)} Molecular-level reference mapping on four gene-property tasks. GenePT, scGPT, scProtoTransformer, and PertMind gene embeddings are evaluated with logistic-regression (LR) and random-forest (RF) classifiers; bars report area under the receiver operating characteristic curve (AUC). \textbf{(b)} Cellular-level perturbation-response prediction on HepG2, K562, and RPE1 cells. Cell representations from scProtoTransformer, STATE-SE, STATE-SM, and PertMind are coupled to a STATE-based expression decoder and evaluated using differential-expression similarity (DES), perturbation-discrimination score (PDS), and mean absolute error (MAE). Higher DES and PDS and lower MAE are better. Error bars denote s.d. over five independent runs. \textbf{(c)} Donor-level tumor-state reference mapping at 50\%, 75\%, and 100\% of the available reference-training donors. Curves report accuracy, AUC, and F1 score; error bars denote s.d. over seven independent runs.}
\label{fig:multiscale-representation}
\end{figure}

GenePT showed that literature-derived descriptions can be converted into gene embeddings and composed with expression measurements to represent single cells \citep{chen2025simple}. We extended this premise to ask whether the biological profiles produced by a perturbation-trained language model provide a reusable representation hierarchy. PertMind first generates a functional profile for each gene, which is encoded with text-embedding-ada-002. A task-specific projection maps the raw text embedding to the 512- or 2,048-dimensional interface required downstream. Expression-weighted aggregation then composes gene embeddings into cell embeddings, and an attention-based multi-instance learning module aggregates cells into donor embeddings. Appendix Figure~\ref{fig:cross-scale-construction} gives the complete construction.

At the molecular level, we evaluated four reference-mapping tasks introduced in the Geneformer benchmark and adopted by scProtoTransformer \citep{tang2025scprototransformer}. These tasks distinguish long- from short-range transcription factors, dosage-sensitive from dosage-insensitive genes, bivalent from non-methylated genes, and bivalent from Lys4-only methylated genes (Figure~\ref{fig:multiscale-representation}a). Across LR and RF classifiers, PertMind embeddings were competitive with the strongest representation in each task and generally approached scProtoTransformer, despite being obtained from generated biological text rather than large-scale single-cell pretraining. The corresponding genome-wide UMAP retained separable gene-biotype structure (Appendix Figure~\ref{fig:gene-embedding-umap}); we treat this visualization as qualitative support rather than an independent performance estimate.

We next tested whether expression-weighted PertMind embeddings preserved sufficient cellular-state information for quantitative perturbation modeling. Using STATE as the expression decoder \citep{adduri2025predicting}, we predicted post-perturbation expression profiles in HepG2, K562, and RPE1 cells (Figure~\ref{fig:multiscale-representation}b). PertMind approached the scProtoTransformer representation across DES, PDS, and MAE, while producing a stronger overall metric profile than the standard STATE-SM representation. Thus, the same text-derived gene space that supported molecular annotation could be composed into cell embeddings capable of resolving continuous perturbation responses, rather than only the discrete \(\mathrm{Up}\), \(\mathrm{Down}\), and \(\mathrm{No}\) outcomes used to train PertMind.

Finally, we aggregated PertMind cell embeddings into donor representations and evaluated tumor-state reference mapping on the pan-cancer setting used by scTransMIL \citep{tang2025sctransmil}. PertMind remained stable as the reference-training set was reduced from all available donors to one half (Figure~\ref{fig:multiscale-representation}c). Its advantage was most apparent in the 50\% regime, whereas scTransMIL improved more strongly as additional labeled donors became available. This behavior indicates that the PertMind representation retained transferable biological structure when donor-level supervision was limited.

Together, these results establish a hierarchical representation route from generated gene knowledge to cells and donors. They do not imply that all three scales occupy an identical latent geometry, because task-specific projections and learned aggregation remain necessary. Within the evaluated tasks, however, a common source of PertMind-derived gene knowledge supported competitive representations at each scale, including settings with limited donor-level supervision.

\section{Discussion}
\label{sec:discussion}

The central finding is that experimentally measured perturbation endpoints can train a reusable biological language-model policy, rather than merely improve a forward-response classifier. PertMind strengthened response prediction in unseen cellular contexts while largely retaining the general capabilities of its pretrained foundation. More importantly, the resulting policy transferred across task direction, perturbation composition, functional interpretation, and representation scale. These outcomes connect a narrowly computable training signal to a broader set of biological operations. The common element is not a shared output format, because the transferred tasks require rankings, process names, mechanistic briefs, or continuous representations. Instead, they require relating interventions, pathways, genes, and cellular states under changing query structures. Perturbation measurements therefore offer more than labels for supervised response models. They provide experimentally anchored consequences that can shape how a language model organizes and deploys biological knowledge. Interpreted as selection pressure, rewards across many contexts may favour integrated strategies linking interventions, pathways, genes, and cellular states. This explanation concerns the observed transfer rather than directly measuring the policy's internal operations.

The results support a selection-and-concentration interpretation of this learning process. The gene-level reward supplies terminal selection pressure by distinguishing trajectories that reach the measured response from those that do not. The pathway reward adds structured intermediate guidance, favoring predictions that remain consistent with coordinated transcriptional changes before the final gene decision. In the evaluated setting, pathway text alone did not reproduce the gain obtained when pathway predictions also received reward. The shuffled-label control further indicates that improvement depends on biological alignment rather than additional reward density or output structure. Pre-RL best@\(N\) provides the complementary starting point: correct conclusions already appeared within the Base model's sampling support. Yet greedy decoding and majority voting failed to concentrate that support into reliable decisions. Reinforcement learning can therefore be understood as reallocating probability toward useful strategies already accessible to the pretrained model. Trusted-trajectory SFT first stabilizes their expression, while GRPO repeatedly selects among them using experimental endpoints and pathway checkpoints. This account explains why post-training can amplify reasoning without requiring exhaustive annotations of every mechanistic step.

Transfer beyond the post-training objective provides the strongest evidence that the selected strategies are reusable. In reverse perturbation-condition prediction, PertMind inferred interventions from observed cell-state transitions, reversing the causal query presented during training. It approached a specialized transition model on primary-cell perturbations \citep{schmidt2022crispr,wang2025cellnavi}, despite receiving no reverse-task adaptation. On unseen double perturbations, PertMind improved the second component's rank and reduced the imbalance between both ground-truth components \citep{norman2019exploring}. This pattern is consistent with compositional reasoning, but does not verify that the model internally composed candidate effects. AssayBench extended transfer from response inference to phenotypic-screen prioritization \citep{de2026assaybench}. Screen-specific PertMind briefs improved different ranking backbones, indicating that the useful output was portable biological context rather than backbone-specific prompting. GeneAgent evaluation added process-level abstraction, where PertMind context improved lexical and semantic agreement with curated biological interpretations \citep{wang2025geneagent}. Collectively, these tasks alter input direction, output type, biological scale, and executive model. Success across those changes is more informative than another in-domain score because none was directly optimized during post-training. Under our operational definition, this pattern supports emergent biological reasoning as transferable task competence.

The representation results reveal a second route by which this competence becomes reusable. PertMind-generated gene profiles externalize distributed model knowledge as natural-language objects that can be encoded once and adapted for distinct biological tasks. This construction extends the GenePT premise that literature-derived descriptions can provide effective molecular representations \citep{chen2025simple}. More broadly, structured intermediate representations have supported modality binding \citep{huang2024tmbl}, semantic-capacity balancing across heterogeneous inputs \citep{xu2026lever}, weakly supervised relational feature mining \citep{tang2023posture}, and pretrained structural priors for molecular generation \citep{tian2026prime}. In single-cell analysis specifically, CoVEL and Modal-NexT integrate heterogeneous modalities by combining complementary views or typed cell--feature relations \citep{tang2024covel,tang2026modalnext}. PertMind addresses a different setting: it uses generated biological language as the common prior and tests whether that prior composes across molecular, cellular, and donor scales. Here, descriptions shaped by perturbation training supported molecular annotation and also composed with expression measurements into cellular representations. Aggregating those cellular representations further produced donor-level embeddings that remained useful when reference supervision was scarce. The hierarchy suggests that language can serve as an interface between symbolic biological knowledge and continuous data models. Functional statements about genes become a representation prior, while measured expression determines how those statements combine within a particular cell. Learned attention then selects cellular evidence relevant to a donor-level state. This role also connects PertMind to the model-as-memory perspective \citep{quek2026memo}. A specialized model can expose internalized domain structure to another predictor through generated context or embeddings, without transferring all parameters. Natural-language knowledge is thus both interpretable communication and a substrate for multiscale representation learning.

Several limitations define the scope of this interpretation. Terminal rewards may admit shortcut strategies that predict the correct endpoint without reconstructing the relevant biological mechanism. Free-text trajectories are auditable, but their individual steps are not causally verified by the reward. Trusted-trajectory filters improve evidential grounding and structure, yet they cannot establish that generated explanations faithfully report the computations producing each answer. Related mechanistic analyses of vision--language models likewise show that apparently plausible outputs can coexist with separable latent circuits that contribute to hallucination \citep{liu2026dualpathway}; this result does not establish the same circuitry in PertMind, but it illustrates why output plausibility alone is insufficient evidence of faithful computation. Pathway supervision is also a transcriptional proxy based on coordinated expression among annotated members. It does not directly measure protein activity, metabolic flux, spatial signaling, or causal pathway activation. The evaluated cellular systems, interventions, datasets, and downstream tasks cover only a limited portion of biological diversity. Generalization to other tissues, disease states, doses, modalities, or perturbation classes consequently remains unresolved. Retrieval resources and model-generated trusted trajectories may propagate omissions, annotation errors, and historical biases from their sources. Such biases could privilege well-studied genes, drugs, and pathways during both initialization and inference. Finally, PertMind-derived embeddings require task-specific projections, learned aggregation, and downstream supervision. Their performance therefore reflects both the language-derived prior and the adapters trained for each benchmark, not a universally sufficient fixed representation.

These findings suggest several directions for developing perturbation-derived reasoning systems. Genetic perturbations could expand the intervention vocabulary and separate target-specific effects from compound pharmacology. Multi-omic readouts could reward consistency across transcription, chromatin, protein, and metabolic responses, providing richer checkpoints along a biological cascade. Spatial assays could test whether reasoning captures interactions among neighboring cells and tissue compartments. Temporal measurements could distinguish early signaling events from delayed compensatory responses and clarify direction changes across response phases. Laboratory feedback loops could turn model proposals into new perturbation experiments, then return measured outcomes as fresh reinforcement signals. Such loops would prioritize informative interventions rather than passively consuming fixed atlases. More direct process supervision could score experimentally supported intermediate events, calibrated uncertainty, and evidence use alongside terminal correctness. Combining these signals with endpoint rewards may improve mechanistic fidelity while preserving scalable supervision. The broader opportunity is to treat expanding perturbation resources as environments for learning how biological systems respond, not only as tables for fitting predictions. Perturbation-derived reinforcement learning can thereby provide a scalable bridge from experimental data to reusable biological reasoning.

\section{Data and Code Availability}
\label{sec:data-code-availability}

The source datasets analysed in this study are available from the original public resources cited in Section~\ref{sec:experimental-setup}. The PertMind project page is available at \url{https://shapsider.github.io/PertMind/}. Source code and usage documentation are available at \url{https://github.com/shapsider/PertMind}. Model weights, tokenizer files, and inference scripts are available at \url{https://huggingface.co/tzcfly/PertMind}.



\bibliography{iclr2027_conference}
\bibliographystyle{iclr2027_conference}

\appendix

\clearpage
\newpage
\section{Additional Methods Details}
\label{app:methods-details}

This appendix collects the reproducibility details deferred from Section~\ref{sec:methods}. Notation, symbols, and labels follow the main text.

\subsection{Data construction details}
\label{app:data-details}

\paragraph{Full experimental condition and drug-level collapsing.} Each Tahoe-100M \texttt{pseudobulk\_differential\_expression} record is indexed by a full experimental condition
\begin{equation}
\tilde d = (d, \mathrm{dose}, \mathrm{plate}),
\end{equation}
where \(d\) is the drug identity, and provides DESeq2-style statistics
\begin{equation}
(\texttt{baseMean}, \texttt{log2FoldChange}, \texttt{lfcSE}, \texttt{stat}, \texttt{pvalue}, \texttt{padj}),
\end{equation}
together with drug, concentration, plate, and cell-line identifiers and the number of treated and control cells. At the condition granularity we write \(\Delta_{c,\tilde d,g} := \texttt{log2FoldChange}\), \(q_{c,\tilde d,g} := \texttt{padj}\), and \(\mu_{c,\tilde d,g} := \texttt{baseMean}\). Records with missing \(\Delta\) or \(q\) are discarded, and to avoid unstable labels from underpowered comparisons we retain only conditions with treated-group cell count \(n^{\mathrm{trt}}_{c,\tilde d} \geq 30\).

The gene-centered task of Section~\ref{sec:problem-formulation} requires one record per \((c, d, g)\) for primary label assignment and confidence weighting. For these two purposes, we merge the surviving conditions \(\tilde d\) that share the drug identity \(d\) by keeping the record with the largest \(|\Delta_{c,\tilde d,g}|\), and write \(\Delta_{c,d,g}\), \(q_{c,d,g}\), \(\mu_{c,d,g}\) for the statistics of the retained record. A separate reward-side condition table retains \(\tilde d\), so pathway-response proxies are computed within individual conditions rather than from these collapsed gene-level records. The collapsing rule makes the primary drug-level label semantics deliberately asymmetric: a directional label asserts that \emph{at least one} tested condition surviving quality control reached the thresholds of Section~\ref{sec:dataset}, whereas a \(\mathrm{No}\) label asserts the complementary statement that no such condition provided directional evidence. Selecting the strongest available effect shifts the drug-level label distribution toward directional calls and raises the typical retained \(|\Delta|\) relative to a dose-averaged summary; it should therefore be interpreted as a maximum-observed-response endpoint rather than a dose-invariant effect.

\paragraph{Confidence weighting.} Each triplet with \(y_g \in \mathcal{Y}\) carries a confidence weight \(w^{\mathrm{conf}}_{c,d,g} \in [0.1, 1]\) computed from \((\Delta, q, \mu) = (\Delta_{c,d,g}, q_{c,d,g}, \mu_{c,d,g})\). For \(\mathrm{Up}\) or \(\mathrm{Down}\),
\begin{equation}
\begin{aligned}
w^{\mathrm{conf}}_{c,d,g} = \max\Big[ 0.1, \ &\operatorname{clip}\!\left(\tfrac{|\Delta|}{2},\, 0,\, 1\right) \\
&\cdot \operatorname{clip}\!\left(\tfrac{-\log_{10}(\max(q,\, 10^{-300}))}{5},\, 0,\, 1\right) \Big],
\end{aligned}
\end{equation}
and for \(\mathrm{No}\),
\begin{equation}
\begin{aligned}
w^{\mathrm{conf}}_{c,d,g} = \max\Big[ 0.1, \ &\operatorname{clip}\!\left(\tfrac{\log_{10}(\mu+1)}{\log_{10}(101)},\, 0,\, 1\right) \\
&\cdot \operatorname{clip}\!\left(\tfrac{0.20-|\Delta|}{0.20},\, 0,\, 1\right) \Big].
\end{aligned}
\end{equation}
These weights give larger effective supervision to strong, significant directional calls and to high-expression, near-zero-effect null calls, while every triplet retains its assigned label. Where the weight enters the RL update is specified in Section~\ref{sec:grpo} and formalized in Appendix~\ref{app:grpo-details}.

\paragraph{Split, class stratification, and leakage.} Before quality control or triplet construction, every Tahoe-100M sample is assigned by cell line. All samples from the five lines listed in Section~\ref{sec:dataset} form the held-out test domain, and all samples from every other cell line form the development domain. After applying the same eligibility and label rules within both domains, development \((c,d,g)\) groups are split 90:10 into training and validation partitions, stratified by cell line and label; records sharing a \((c,d,g)\) key cannot cross this boundary. Because the outer partition is over cell lines rather than drugs or genes, a drug or gene may occur in both development and test domains in different cell contexts. A supporting case may share the current query's \((d,g)\) pair only when it comes from the training partition and is then treated as cross-context evidence. The current query, its repeated measurements, validation outcomes, and all held-out-cell outcomes are excluded from retrieval. Single-cell replicates are already aggregated in the Tahoe-100M pseudobulk statistics, so each candidate example corresponds to one eligible cell-line--perturbation--gene record. We retain every eligible triplet and use cell-line-, drug-, and label-stratified batches and prompt streams to control optimization imbalance; no class, cell line, or drug is capped or downsampled. The leakage controls in Section~\ref{sec:dataset} apply throughout: no query-specific expression measurement enters the model side, and pathway supervision is available only on the reward side.

\subsection{Trajectory and SFT details}
\label{app:sft-details}

\paragraph{Retrieval configuration.} The retrieval module aligns entities from PubChem, DrugBank, UniProt, Gene Ontology, Reactome, STRING, and CORUM by standard identifiers, filters high-frequency low-information nodes and duplicate facts before prompt construction, and computes drug and gene similarity from semantic annotation and knowledge-graph topology, restricted to \(\mathcal{C}_{\mathrm{train}}\). Cell-context similarity is computed only from outcome-independent tissue, lineage, and curated genotype annotations; no perturbation-response-derived expression feature is used for retrieval or ranking. Contrastive cases are retrieved separately from each outcome group and united as \(\mathcal{S}(x) = \mathcal{S}_{\mathrm{Up}}(x) \cup \mathcal{S}_{\mathrm{Down}}(x) \cup \mathcal{S}_{\mathrm{No}}(x)\). A training case from another cell line may share the query's \((d,g)\) pair and is then included only as cross-context evidence; the current query, its repeated measurements, and every held-out-cell outcome are excluded. The assembled context is
\begin{equation}
\mathcal{K}(x) = \big(\text{knowledge-graph evidence for } (c, d, g), \ \mathcal{S}(x)\big).
\end{equation}
No expression measurement or label of the current query enters \(\mathcal{K}(x)\), so inference relies solely on knowledge-graph evidence and outcome-stratified training cases with their historical outcomes. Figure~\ref{fig:prompt-schema} summarizes the resulting model-visible prompt fields, required response order, and reward-side quantities that are withheld from the prompt.

\begin{figure}[tbp]
\centering
\begingroup
\setlength{\fboxsep}{3pt}
\setlength{\fboxrule}{0.45pt}
\footnotesize
\begin{minipage}[t]{0.55\linewidth}
\fcolorbox{black}{blue!4}{%
\begin{minipage}[t]{0.99\linewidth}
\textbf{INPUT: model-visible prompt context}
\smallskip

\textbf{Task/query.} Infer the response of target gene \(g\) under cell line \(c\) and drug \(d\); output vocabulary is \(\mathrm{Up}/\mathrm{Down}/\mathrm{No}\).
\smallskip

\textbf{KG/static context.} Drug mechanism and known targets; target-gene function and regulatory context; curated drug--target and pathway relations with source identifiers.
\smallskip

\textbf{Training supporting cases.} Examples drawn only from \(\mathcal{C}_{\mathrm{train}}\), with source triplets and historical outcomes; same-\((d,g)\) cases from other training cell lines are cross-context evidence, while the current query, repeated measurements, and held-out-cell outcomes are excluded.
\smallskip

\textbf{Complete candidate pathways.} Full \(\mathcal{P}^{\mathrm{cand}}(d,g)\); each entry gives pathway name, documented relation, and representative non-target members selected by the pre-registered outcome-independent static score.
\smallskip

\textbf{Hidden boundary.} Current-query labels, current-query DE statistics, pathway proxy labels, determinacy masks, and rewards are not visible to the model.
\end{minipage}}%
\end{minipage}%
\hfill
\begin{minipage}[t]{0.42\linewidth}
\fcolorbox{black}{green!5}{%
\begin{minipage}[t]{0.94\linewidth}
\textbf{OUTPUT: required response order}
\smallskip

\texttt{pathway\_directions}: one \(\mathrm{Up}\), \(\mathrm{Down}\), or \(\mathrm{No}\) prediction for every supplied pathway in \(\mathcal{P}^{\mathrm{cand}}(d,g)\).
\smallskip

\texttt{mechanistic\_explanation}: concise evidence-grounded drug-target--pathway--gene chain; unsupported statements must be marked as inference.
\smallskip

\texttt{Final Answer}: exactly one target-gene label, \(\mathrm{Up}\), \(\mathrm{Down}\), or \(\mathrm{No}\). No \texttt{uncertain} class.
\end{minipage}}%

\vspace{0.45em}
\fcolorbox{black}{orange!8}{%
\begin{minipage}[t]{0.94\linewidth}
\textbf{REWARD: direct response scoring}
\smallskip

\(R_{\mathrm{gene}}\): compare \(\hat y_{g,i}\) with \(y_g\).
\smallskip

\(R_{\mathrm{pw}}\): compare model outputs
\(\{\hat y_{P,i}:P\in\mathcal{P}^{\mathrm{det}}(c,d,g)\}\) with
\(\{y_P(c,d,g):P\in\mathcal{P}^{\mathrm{det}}(c,d,g)\}\);
the model declares all candidates, but only this determinate subset is scored.
\smallskip

\(R_{\mathrm{fmt}}\): validate required fields, vocabulary, and one terminal answer.
\begin{equation*}
R=R_{\mathrm{gene}}+\lambda_{\mathrm{pw}}m_{\mathrm{pw}}R_{\mathrm{pw}}
+\lambda_{\mathrm{fmt}}R_{\mathrm{fmt}}.
\end{equation*}
\end{minipage}}%
\end{minipage}
\endgroup
\caption{Prompt and reward interface schema for PertMind trajectory generation. The left panel shows model-visible input, the upper-right panel shows the generated output schema, and the lower-right panel shows direct reward-side scoring. Condition-level pathway quantities are preprocessing intermediates, not direct response-scoring inputs, and are therefore omitted.}
\label{fig:prompt-schema}
\end{figure}

\paragraph{Model-generated trajectory sampling.} Given \(\mathcal{K}(x)\), the current Qwen3-4B policy generates \(M\) candidate trajectories,
\begin{equation}
\tilde z_m, \hat y_m \sim \pi_{\theta_0}(\cdot \mid x, \mathcal{K}(x)), \qquad m = 1, \dots, M,
\end{equation}
where \(\theta_0\) denotes the Qwen3-4B Base parameters before any PertMind-specific update, rather than an external teacher, and \(M=8\). The model is prompted to (1) identify the perturbation's direct targets and mechanism of action, (2) compare retrieved cases to the current drug, gene, and cell context, (3) predict a direction for every supplied candidate pathway, (4) propagate pathway-level change to the regulation of the target gene, and (5) output a single \(\mathrm{Up}\), \(\mathrm{Down}\), or \(\mathrm{No}\) conclusion. Trusted-trajectory checks are implemented as reproducible validators: deterministic parsers verify the unique terminal label, required pathway slots, and output order; every factual entity--relation claim must cite an identifier present in the prompt-visible evidence, while uncited statements must be marked explicitly as inference; and required structured fields verify the presence of a perturbation-target--pathway--gene chain. Together with final-label correctness, these checks instantiate the five criteria of Section~\ref{sec:trajectory}. Retained trajectories form \(\mathcal{D}_{\mathrm{traj}} = \{(x, z^\star, y_g)\}\).

\paragraph{Balanced Fine-Tuning objective.} For sample \(b\) with reference response tokens \(y^\star_b\), let
\begin{equation}
c_{b,t} = \pi_\theta(y^\star_{b,t} \mid y^\star_{b,<t}, x_b), \qquad \ell_{b,t} = -\log c_{b,t}.
\end{equation}
Using a local window of \(g_{\mathrm{BFT}} = 256\) response tokens around \(t\), let \(W_{b,t}\) denote the valid response-token indices in that window after clipping it at sequence boundaries, and define the local context confidence
\begin{equation}
C^{\mathrm{loc}}_{b,t} = \frac{1}{|W_{b,t}|} \sum_{j \in W_{b,t}} c_{b,j},
\end{equation}
the locally normalized token weight
\begin{equation}
\rho_{b,t} = \operatorname{clip}\!\left(\frac{c_{b,t}}{C^{\mathrm{loc}}_{b,t} + \epsilon},\, 0,\, 1\right), \qquad w^{\mathrm{BFT}}_{b,t} = \operatorname{sg}(\rho_{b,t}),
\end{equation}
and the sequence-level difficulty weight
\begin{equation}
p^{\mathrm{conf}}_b = \min_t C^{\mathrm{loc}}_{b,t}, \qquad s_b = \operatorname{sg}(1 - p^{\mathrm{conf}}_b),
\end{equation}
where \(\operatorname{sg}\) denotes stop-gradient. The BFT objective, used as the SFT loss on \(\mathcal{D}_{\mathrm{traj}}\), is
\begin{equation}
\mathcal{L}_{\mathrm{SFT}}(\theta) = \frac{1}{B}\sum_{b=1}^{B} s_b \frac{\sum_t m_{b,t} w^{\mathrm{BFT}}_{b,t} \ell_{b,t}}{\sum_t m_{b,t} + \epsilon},
\end{equation}
where \(m_{b,t}\) masks non-response tokens and \(\epsilon>0\) is a numerical stabilizer. Isolated low-confidence tokens embedded in high-confidence contexts are down-weighted; because \(s_b\) depends on the minimum local confidence, trajectories containing at least one difficult local span receive larger relative sequence-level weights than easier trajectories. All confidence-derived quantities carry a stop-gradient so that the model cannot lower the loss by manipulating weights instead of raising the likelihood of \(y^\star\). One epoch on \(\mathcal{D}_{\mathrm{traj}}\) produces the SFT policy \(\pi_{\theta_{\mathrm{SFT}}}\), which is retained as both the GRPO initialization and the fixed KL reference in Appendix~\ref{app:grpo-details}.

\subsection{Pathway supervision details}
\label{app:pathway-details}

\paragraph{Candidate pathway selection.} We collect Reactome pathways containing the target gene,
\begin{equation}
\mathcal{P}_0(g) = \{P : g \in P\},
\end{equation}
and retain by default only pathways with \(10 \leq |P| \leq 200\) members. For each candidate the non-target member set is \(P^{-g} = P \setminus \{g\}\); the target gene is excluded from every subsequent pathway score so that its own expression cannot leak into the auxiliary reward. Candidates are ranked using only information that is available before any \((c, d)\)-specific expression outcome is examined,
\begin{equation}
R_{\mathrm{cand}}(P \mid d, g) = \alpha_{\mathrm{gene}} R^{\mathrm{cand}}_{\mathrm{gene}}(P \mid g) + \alpha_{\mathrm{drug}} R^{\mathrm{cand}}_{\mathrm{drug}}(P \mid d),
\end{equation}
where \(R^{\mathrm{cand}}_{\mathrm{gene}}(P \mid g)\) scores the curated strength or confidence of target-gene membership in the pathway and \(R^{\mathrm{cand}}_{\mathrm{drug}}(P \mid d)\) quantifies drug--pathway relevance. We use \(\alpha_{\mathrm{gene}}=0.60\) and \(\alpha_{\mathrm{drug}}=0.40\). Drug relevance is prioritized in the following order: a direct drug target within the pathway; a target within one or two hops of pathway members in a high-confidence interaction network; or shared membership on a manually curated mechanism-of-action axis. When no reliable drug-mechanism annotation is available, this term is recorded as unknown rather than inferred speculatively. The top-ranked candidates are kept, at most 3 per query, as \(\mathcal{P}^{\mathrm{cand}}(d,g)\). Both the filter and the ranking are computed before the \((c, d)\)-specific differential-expression outcome is examined and depend only on curated target/pathway membership and drug-target or drug--pathway knowledge, so candidate selection is reproducible at inference time and outcome-independent.

\paragraph{Condition-level pathway-response proxy.} Label construction may use training-split expression statistics, but the resulting label is used only for reward computation, never as model input. Pathway scores are always computed inside a single complete experimental condition so that every member entering a score shares the same dose and plate. Let \(\mathcal{T}(c, d)\) denote the set of quality-controlled conditions \(\tilde d = (d, \mathrm{dose}, \mathrm{plate})\) for drug \(d\) in cell line \(c\) that survive the filters of Appendix~\ref{app:data-details}. For each \(P \in \mathcal{P}^{\mathrm{cand}}(d,g)\) and each \(\tilde d \in \mathcal{T}(c, d)\), let \(H(P, c, \tilde d, g) \subseteq P^{-g}\) denote the non-target members with finite \(\Delta_{c,\tilde d,h}\) and \(q_{c,\tilde d,h}\) under a condition satisfying \(n^{\mathrm{trt}}_{c,\tilde d}\geq30\). The entire set is drawn from a single \((c, \tilde d)\), so \(H\) never mixes members measured at different doses or on different plates. For \(h \in H(P, c, \tilde d, g)\), define the condition-level significance-weighted signed effect
\begin{equation}
u_{h, \tilde d} = \Delta_{c, \tilde d, h} \cdot \operatorname{clip}\!\left(-\log_{10}(\max(q_{c, \tilde d, h}, q_{\min})), 0, a_{\max}\right),
\end{equation}
where \(q_{\min}=10^{-6}\) is a fixed lower bound and \(a_{\max}=6\) caps the influence of any single, extremely small \(q\)-value. The condition-level pathway aggregate score is
\begin{equation}
S_{P, \tilde d} = \frac{\sum_{h \in H(P, c, \tilde d, g)} w_h \, u_{h, \tilde d}}{\sum_{h \in H(P, c, \tilde d, g)} w_h},
\end{equation}
with \(w_h = 1\) by default; non-uniform, outcome-independent topological weights are used only when pre-registered. The condition-level fraction of nonzero-effect members whose sign agrees with the aggregate direction is
\begin{equation}
C_{P, \tilde d} = \frac{|\{h \in H(P, c, \tilde d, g) : \operatorname{sign}(u_{h, \tilde d}) = \operatorname{sign}(S_{P, \tilde d}), \ u_{h, \tilde d} \neq 0\}|}{|\{h \in H(P, c, \tilde d, g) : u_{h, \tilde d} \neq 0\}|},
\end{equation}
with the convention \(C_{P, \tilde d} = 0\) when no member has nonzero effect. The condition-level fraction of significantly responding members is
\begin{equation}
A_{P, \tilde d} = \frac{|\{h \in H(P, c, \tilde d, g) : q_{c, \tilde d, h} \leq 0.05 \ \land \ |\Delta_{c, \tilde d, h}| \geq 0.585\}|}{|H(P, c, \tilde d, g)|}.
\end{equation}
The condition-level pathway-response proxy label is
\begin{equation}
y_{P, \tilde d} =
\begin{cases}
\mathrm{uncertain}, & |H(P, c, \tilde d, g)| < 5, \\
\mathrm{Up}, & S_{P, \tilde d} \geq \tau_{\mathrm{dir}} \ \land \ C_{P, \tilde d} \geq \gamma, \\
\mathrm{Down}, & S_{P, \tilde d} \leq -\tau_{\mathrm{dir}} \ \land \ C_{P, \tilde d} \geq \gamma, \\
\mathrm{No}, & |S_{P, \tilde d}| \leq \tau_{\mathrm{no}} \ \land \ A_{P, \tilde d} \leq \eta_{\mathrm{no}}, \\
\mathrm{uncertain}, & \text{otherwise},
\end{cases}
\end{equation}
with \(\gamma=0.60\), \(\tau_{\mathrm{dir}}=0.20\), \(\tau_{\mathrm{no}}=0.05\), and \(\eta_{\mathrm{no}}=0.20\). The directional and null cutoffs are constrained to satisfy
\begin{equation}
0 \leq \tau_{\mathrm{no}} < \tau_{\mathrm{dir}},
\end{equation}
so the directional region \(|S_{P, \tilde d}| \geq \tau_{\mathrm{dir}}\) and the null region \(|S_{P, \tilde d}| \leq \tau_{\mathrm{no}}\) are disjoint. All parameters are fixed on the development split before evaluation on the held-out cell lines. Requiring both a small aggregate effect and a low fraction of significantly responding members for \(\mathrm{No}\) prevents mislabeling a pathway with large but mutually canceling up- and down-regulated members as unresponsive. Conditions with fewer than five reliable non-target members or with effects in the gray zone are labeled \(\mathrm{uncertain}\) at the condition level. We refer to \(y_{P, \tilde d}\) as a transcriptional pathway-response proxy: it summarizes the coordinated transcriptional response of the pathway's non-target members under a single \((c, \tilde d)\) and is not a direct measurement of protein activity, metabolic flux, or causal pathway activation.

\paragraph{Cross-condition consensus for drug-level pathway labels.} The drug-level label \(y_P(c, d, g)\) used by the reward in Section~\ref{sec:grpo} is obtained by a conservative consensus over the determinate condition-level labels in \(\{y_{P, \tilde d} : \tilde d \in \mathcal{T}(c, d)\}\), so that no drug-level directional or null call is issued when determinate conditions disagree. Conditions labeled \(\mathrm{uncertain}\) are ignored by the consensus, and a single determinate condition can supply the drug-level label. Let
\begin{equation}
\mathcal{L}_{P, c, d, g} = \{y_{P, \tilde d} : \tilde d \in \mathcal{T}(c, d), \ y_{P, \tilde d} \neq \mathrm{uncertain}\}
\end{equation}
be the set of determinate condition-level labels. Then
\begin{equation}
y_P(c, d, g) =
\begin{cases}
\mathrm{uncertain}, & \mathcal{L}_{P, c, d, g} = \emptyset, \\
\mathrm{Up}, & \mathcal{L}_{P, c, d, g} \neq \emptyset \ \land \ \mathcal{L}_{P, c, d, g} \subseteq \{\mathrm{Up}\}, \\
\mathrm{Down}, & \mathcal{L}_{P, c, d, g} \neq \emptyset \ \land \ \mathcal{L}_{P, c, d, g} \subseteq \{\mathrm{Down}\}, \\
\mathrm{No}, & \mathcal{L}_{P, c, d, g} \neq \emptyset \ \land \ \mathcal{L}_{P, c, d, g} \subseteq \{\mathrm{No}\}, \\
\mathrm{uncertain}, & \text{otherwise.}
\end{cases}
\end{equation}
A drug with a single determinate condition inherits that condition's label; a drug with multiple determinate conditions receives a directional or null drug-level label only when all determinate condition-level labels agree, and any disagreement among those determinate labels is mapped to \(\mathrm{uncertain}\). Because the consensus is taken over pre-computed condition-level labels, no drug-level pathway score ever fuses members measured at different doses or on different plates. \(\mathrm{uncertain}\) triplets remain in the main-task training pool and are simply excluded from the pathway reward term.

\paragraph{Model-visible information and pathway masking.} The prompt lists the complete retained candidate set \(\mathcal{P}^{\mathrm{cand}}(d,g)\); for each listed candidate it includes the pathway name, its documented relation to the drug and target gene, and a small number of representative non-target members selected once per pathway using the same pre-registered static score combining annotation confidence and outcome-independent network centrality. This fixed member selection never consults the current condition's expression statistics. The prompt never includes the pathway's proxy direction, the expression effect, \(p\)-value, or \(q\)-value of any member under any condition, the intermediate quantities \(S_{P, \tilde d}\), \(C_{P, \tilde d}\), or \(A_{P, \tilde d}\), or any reordering of candidates based on the realized outcome. In particular, the prompt does not indicate which candidates received a determinate \(y_P(c, d, g)\). The scored subset used in Section~\ref{sec:grpo} is
\begin{equation}
\mathcal{P}^{\mathrm{det}}(c,d,g) = \{P \in \mathcal{P}^{\mathrm{cand}}(d,g) : y_P(c, d, g) \neq \mathrm{uncertain}\}, \qquad K = \big|\mathcal{P}^{\mathrm{det}}(c,d,g)\big|,
\end{equation}
and the pathway mask
\begin{equation}
m_{\mathrm{pw}} = \mathbf{1}[K > 0] \in \{0, 1\}
\end{equation}
is redundant with the \(R_{\mathrm{pw}} := 0\) convention when \(K = 0\); it is kept explicit because the training code applies it as a per-sample mask when reducing rewards over a batch. Triplets whose candidates all turn out \(\mathrm{uncertain}\) remain in the main-task training pool and are simply excluded from the pathway reward term, so pathway augmentation increases supervision density without restricting the training corpus.

\subsection{Full optimization definition}
\label{app:grpo-details}

\paragraph{Grouped rollouts.} GRPO trains on every triplet in \(\mathcal{D}_{\mathrm{train}}\) with a determinate primary label, not only the trusted-trajectory subset used for SFT. For each prompt \(x\), a group of \(G=8\) responses \(o_i\), \(i = 1, \dots, G\), is sampled from the current policy conditioned on the same biological context \(\mathcal{K}(x)\). Each response contains structured pathway predictions, a mechanistic explanation, and a final gene-level label, and is scored by \(R(o_i)\) as in Section~\ref{sec:grpo}. The auxiliary weights are \(\lambda_{\mathrm{pw}}=0.15\) and \(\lambda_{\mathrm{fmt}}=0.05\), selected on the held-out validation subset of the development domain and fixed before evaluation on \(\mathcal{C}_{\mathrm{test}}\).

\paragraph{Group standardization and confidence scaling.} The standardized advantage
\begin{equation}
\hat A_i = \frac{R(o_i) - \operatorname{mean}_j R(o_j)}{\operatorname{std}_j R(o_j) + \epsilon}
\end{equation}
gives zero-variance groups a vanishing gradient contribution; such groups are uninformative rather than harmful. When the primary term alone is uniform across a group, it does not affect the within-group ranking, so the auxiliary terms provide the remaining preference signal. Confidence is applied after group standardization as a positive per-triplet multiplier on the complete sample objective, and never as an additional scaling of \(\hat A_i\) itself, so that the within-group ranking induced by \(\{\hat A_i\}_{i=1}^{G}\) is unchanged and the shared positive constant \(w^{\mathrm{conf}}_{c,d,g}\) enters below at the objective level to uniformly scale the entire parameter update contributed by that triplet.

\paragraph{Clipped, KL-regularized policy optimization.} The behavior policy \(\pi_{\theta_{\mathrm{old}}}\) that generated the current rollouts is refreshed at every rollout iteration and is used only to define the clipped importance ratio below; it is distinct from the frozen SFT reference \(\pi_{\theta_{\mathrm{SFT}}}\), which enters the objective only through the KL penalty. Let the token-level importance ratio against the behavior policy be
\begin{equation}
r_{i,t}(\theta) = \frac{\pi_\theta(o_{i,t} \mid x, \mathcal{K}(x), o_{i,<t})}{\pi_{\theta_{\mathrm{old}}}(o_{i,t} \mid x, \mathcal{K}(x), o_{i,<t})},
\end{equation}
and, separately, let
\begin{equation}
\kappa_{i,t}(\theta) = \frac{\pi_{\theta_{\mathrm{SFT}}}(o_{i,t} \mid x, \mathcal{K}(x), o_{i,<t})}{\pi_\theta(o_{i,t} \mid x, \mathcal{K}(x), o_{i,<t})}
\end{equation}
denote a sampled-token ratio of the frozen SFT reference to the current policy, evaluated only at the sampled token \(o_{i,t}\). The symbol \(\kappa_{i,t}\) is deliberately distinct from the behavior ratio \(r_{i,t}\): its denominator is \(\pi_\theta\) rather than \(\pi_{\theta_{\mathrm{old}}}\), and it drives the KL penalty rather than the clipped policy surrogate, so \(\kappa_{i,t}\) must not be identified with \(r_{i,t}\). This confidence-weighted, KL-regularized GRPO variant \citep{guo2025deepseek} maximizes
\begin{equation}
\mathcal{J}_{\mathrm{GRPO}}(\theta) = \mathbb{E}\!\left[ \frac{1}{G} \sum_{i=1}^{G} w^{\mathrm{conf}}_{c,d,g} \, \frac{1}{|o_i|} \sum_t \left( J^{\mathrm{clip}}_{i,t}(\theta) - \beta \, \mathrm{KL}_{i,t}(\theta) \right) \right],
\end{equation}
so that the per-triplet weight \(w^{\mathrm{conf}}_{c,d,g}\) multiplies the complete per-response objective, including both the clipped policy surrogate and its KL penalty, exactly once. It uniformly scales every parameter update contributed by that triplet without appearing inside \(\hat A_i\) or \(\mathrm{KL}_{i,t}\). The token-level clipped surrogate uses the unscaled standardized advantage,
\begin{equation}
J^{\mathrm{clip}}_{i,t}(\theta) = \min\!\big( r_{i,t}(\theta) \, \hat A_i, \ \operatorname{clip}(r_{i,t}(\theta),\, 1-\varepsilon,\, 1+\varepsilon) \, \hat A_i \big),
\end{equation}
and the token-level KL term uses the sampled-token GRPO estimator
\begin{equation}
\mathrm{KL}_{i,t}(\theta) = \kappa_{i,t}(\theta) - \log \kappa_{i,t}(\theta) - 1,
\end{equation}
which is nonnegative for every \(\kappa_{i,t}(\theta) > 0\) and equals zero exactly when \(\pi_\theta(o_{i,t} \mid \cdot) = \pi_{\theta_{\mathrm{SFT}}}(o_{i,t} \mid \cdot)\); it estimates the KL divergence from the current policy toward \(\pi_{\theta_{\mathrm{SFT}}}\) using only the sampled token \(o_{i,t}\) and never enumerates the full token vocabulary. The clipping term limits the size of policy updates driven by any single batch of rewards, and the KL term toward \(\pi_{\theta_{\mathrm{SFT}}}\), rather than the continually updated policy, preserves the linguistic competence and reasoning structure established during SFT. We use clipping range \(\varepsilon=0.20\), KL coefficient \(\beta=0.01\), and the remaining optimization settings listed in Section~\ref{sec:experimental-setup}.

\paragraph{Training monitors.} Throughout training we monitor gene-level accuracy, per-class recall for \(\mathrm{Up}\), \(\mathrm{Down}\), and \(\mathrm{No}\), pathway-label accuracy on \(\mathcal{P}^{\mathrm{det}}\), output-format pass rate, response length, and policy KL divergence toward \(\pi_{\theta_{\mathrm{SFT}}}\), so that reward-hacking behaviors such as optimizing format alone, over-predicting majority classes, or gaining indirect advantage through response length rather than correctness can be identified and used to trigger early stopping or hyperparameter adjustment on the training-split validation subset.

\subsection{Ablation-variant definitions}
\label{app:ablation-variants}

The six variants in Figure~\ref{fig:results-overview}b share the same Qwen3-4B Base checkpoint, held-out-cell evaluation protocol, decoding configuration, and training/validation partitions. They differ only in post-training initialization, model-visible pathway context, and active reward terms:

\begin{itemize}
    \item \textbf{Base} uses the unmodified Qwen3-4B Base model and receives neither PertMind SFT nor GRPO.
    \item \textbf{SFT} applies the one-epoch BFT-weighted cold start on \(\mathcal{D}_{\mathrm{traj}}\) and receives no subsequent GRPO update.
    \item \textbf{Shuffled pathway} starts from the same SFT policy and uses the full structured pathway prompt and reward interface, but permutes determinate pathway labels within the training partition while preserving label frequency. The gene and format rewards remain correctly aligned. This control tests whether pathway gains require biological alignment rather than merely extra reward density or output structure.
    \item \textbf{Gene-only GRPO} starts from the SFT policy and uses \(R_{\mathrm{gene}}\) as its only biological reward. The format check remains active as a parsing constraint, while pathway candidates and \(R_{\mathrm{pw}}\) are omitted.
    \item \textbf{Prompt-only pathway context} starts from the SFT policy and exposes the same candidate-pathway descriptions and structured pathway slots used by PertMind, but masks \(R_{\mathrm{pw}}\). Its biological learning signal is therefore the same terminal gene-level reward as in gene-only GRPO; the comparison isolates the effect of pathway text without pathway supervision.
    \item \textbf{PertMind} starts from the SFT policy and uses the complete model-visible pathway context together with \(R_{\mathrm{gene}}\), masked \(R_{\mathrm{pw}}\), and \(R_{\mathrm{fmt}}\) as defined in Section~\ref{sec:grpo}.
\end{itemize}

All GRPO-based variants use the same rollout group size, optimizer, confidence weighting, clipping range, and KL coefficient. When a pathway term is absent or masked, it contributes exactly zero to both the reward and the group-standardized advantage.

\subsection{Cross-scale embedding construction}
\label{app:cross-scale-embeddings}

\begin{figure}[t]
\begin{center}
\includegraphics[width=\linewidth]{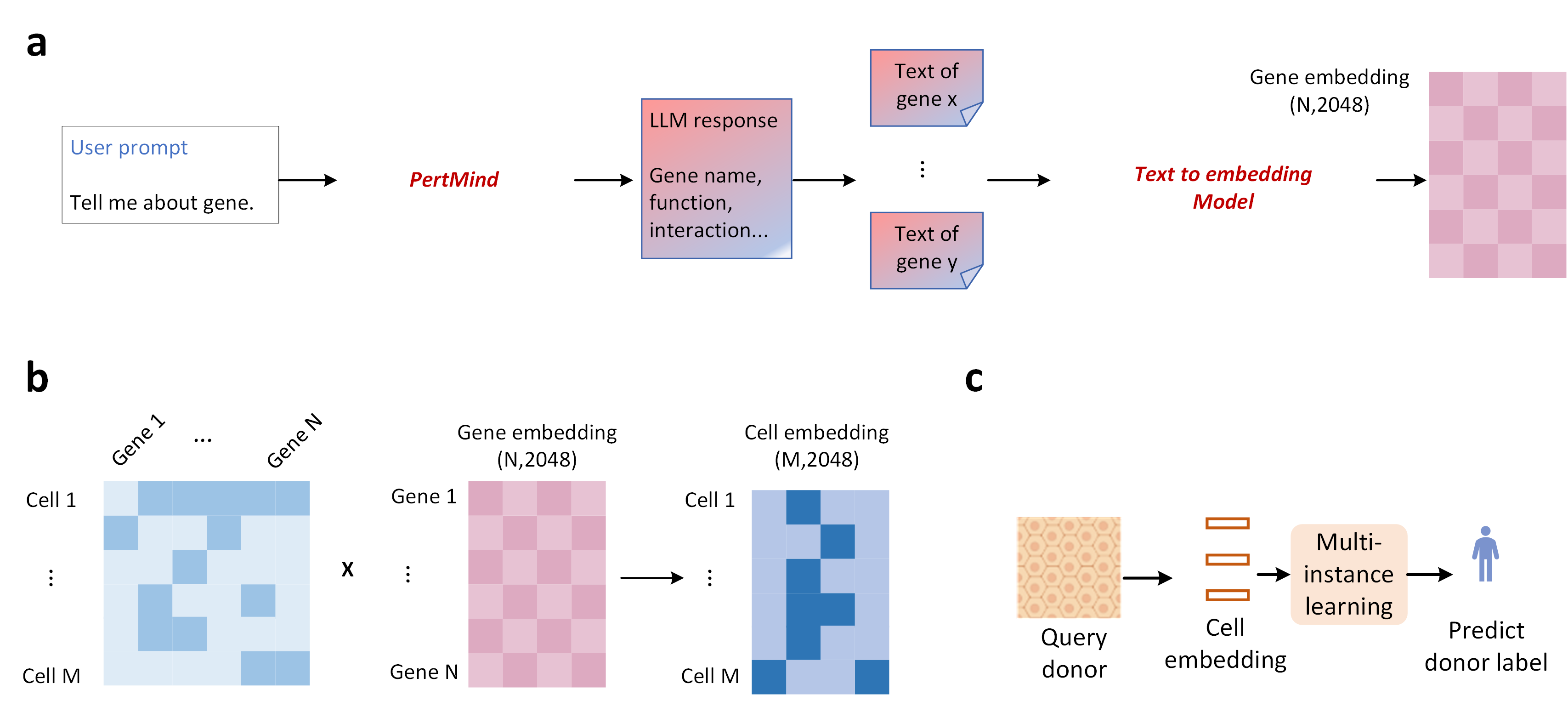}
\end{center}
\caption{Hierarchical construction of PertMind-derived representations. \textbf{(a)} PertMind generates a functional profile for each gene; text-embedding-ada-002 encodes the profile, and a task-specific linear projection adapts the raw 1,536-dimensional text vector to the required 512- or 2,048-dimensional downstream interface. Dimensionality labels in the schematic denote the illustrated 2,048-dimensional post-projection branch. \textbf{(b)} Expression-weighted aggregation composes projected gene embeddings into cell embeddings. \textbf{(c)} Attention-based multi-instance learning aggregates cells from one donor into a donor embedding and predicts the donor label.}
\label{fig:cross-scale-construction}
\end{figure}

Following GenePT \citep{chen2025simple}, we treat generated text as an explicit interface between biological knowledge and numerical representation. For each gene \(g\), PertMind produces a profile \(p_g\) describing its function, interactions, pathways, and context-dependent roles. The raw text embedding and its task-adapted representation are
\begin{equation}
u_g = E_{\mathrm{ada}}(p_g) \in \mathbb{R}^{1536},
\qquad
e_g^{(t)} = A_t u_g + b_t \in \mathbb{R}^{d_t},
\qquad
d_t \in \{512, 2048\},
\end{equation}
where \(E_{\mathrm{ada}}\) denotes text-embedding-ada-002 and \(A_t,b_t\) form the learned projection for downstream task \(t\). The projection permits different decoders to consume a common source representation without claiming that every task shares an identical fixed-dimensional latent space.
Whenever a downstream interface requires this adapter, it is reinitialized within each training fold and optimized jointly with the downstream classifier or decoder using only the reference-training partition; query or test labels are never used to fit the projection.

For cell \(i\), let \(x_{ig}\geq 0\) be the preprocessed expression of gene \(g\). We obtain an expression-weighted cell embedding
\begin{equation}
z_i^{(t)}
=
\operatorname{Normalize}\!\left(
\frac{\sum_{g \in \mathcal{G}_i} x_{ig} e_g^{(t)}}
{\sum_{g \in \mathcal{G}_i} x_{ig} + \epsilon}
\right),
\end{equation}
where \(\mathcal{G}_i\) is the set of represented genes, \(\epsilon>0\) prevents division by zero, and \(\operatorname{Normalize}\) denotes row-wise unit \(\ell_2\) normalization. For donor \(d\), the cells \(\mathcal{I}_d\) form a multi-instance bag. An attention module computes \(\alpha_{di}=\operatorname{softmax}_{i\in\mathcal{I}_d}(a(z_i^{(t)}))\) and aggregates
\begin{equation}
h_d^{(t)} = \sum_{i \in \mathcal{I}_d} \alpha_{di} z_i^{(t)}.
\end{equation}
The resulting donor embedding is passed to the donor-level classifier. Figure~\ref{fig:cross-scale-construction} summarizes this gene-to-cell-to-donor hierarchy.

\clearpage
\newpage
\section{Additional Results}
\label{app:additional-results}

This section collects the pre-RL sampling analysis, gene-embedding visualization, per-subject general evaluation, and literature-audited case deferred from the main Results.

Before reinforcement learning, we probed the Base model's sampling distribution on the forward perturbation task by drawing responses independently for each query up to \(N=50\) rollouts. Here, \emph{oracle} denotes an evaluation-only rule that uses the gold label to determine whether any sampled response is correct; the model cannot implement this selection at inference time. For a fixed query, best@\(N\) counts the query as correct if any of its first \(N\) sampled responses matches the gold terminal label, whereas majority@\(N\) uses the modal parsed terminal label among those \(N\) responses. We additionally report \(\Delta\)best@\(N\), the incremental oracle gain contributed by the \(N\)-th rollout, together with the mean parse rate, the any-parse rate, and class-conditional best@\(N\) for \(\mathrm{No}\), \(\mathrm{Up}\), and \(\mathrm{Down}\).
\label{app:pre-rl-analysis}

\begin{figure}[t]
\begin{center}
\includegraphics[width=\linewidth]{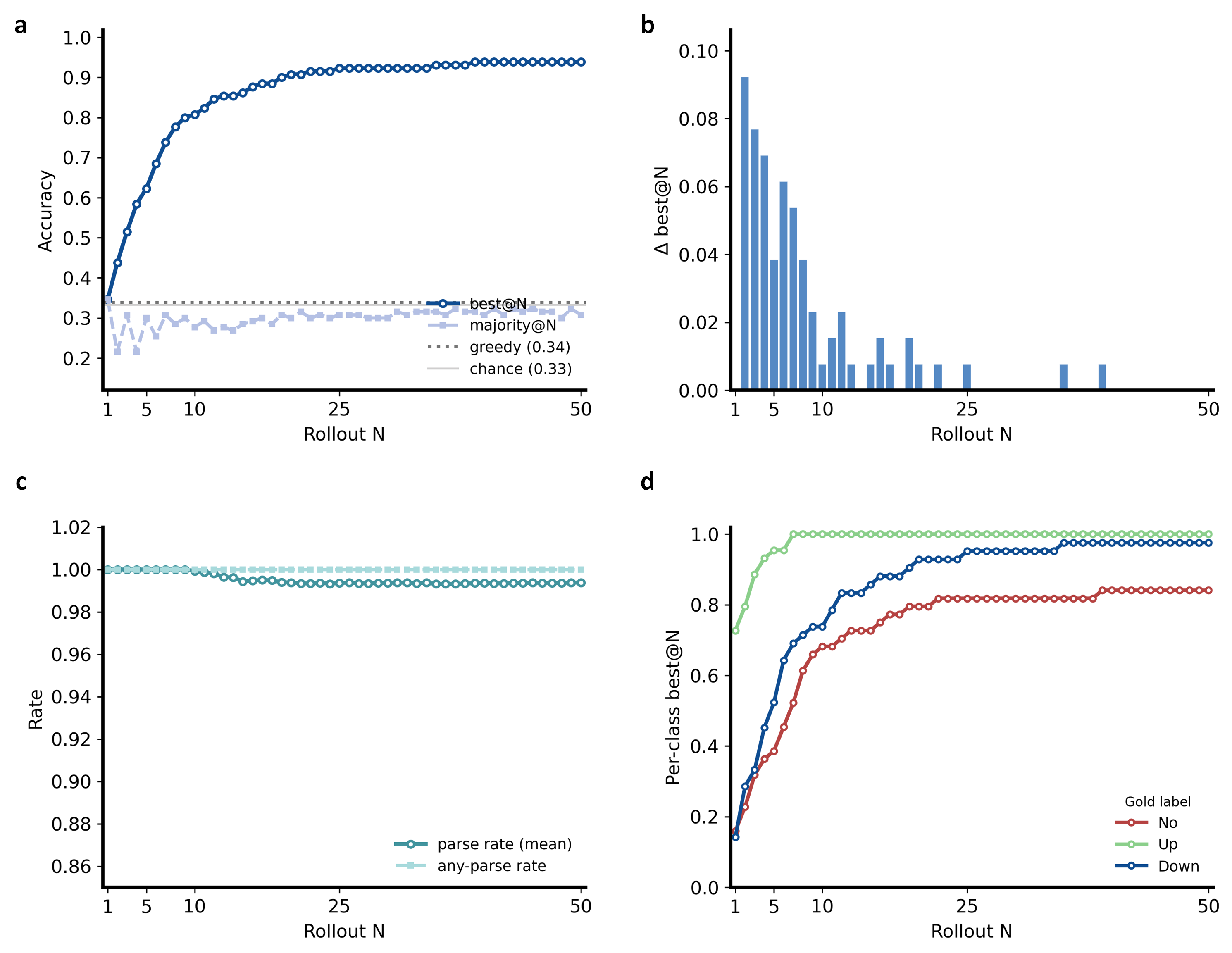}
\end{center}
\caption{Pre-RL sampling analysis of Qwen3-4B Base on the forward perturbation task. \textbf{(a)} Oracle best@\(N\), majority-vote accuracy, greedy decoding, and one-third chance accuracy as the number of sampled rollouts increases. \textbf{(b)} Incremental improvement in best@\(N\) from each additional rollout. \textbf{(c)} Mean parse rate and the probability that at least one of \(N\) responses contains a valid terminal \(\mathrm{No}\), \(\mathrm{Up}\), or \(\mathrm{Down}\) label. \textbf{(d)} Class-conditional best@\(N\) for the three gold labels. Best@\(N\) is an oracle diagnostic of support in the sampling distribution and is not an inference-time selection method.}
\label{fig:pre-rl-analysis}
\end{figure}

Figure~\ref{fig:pre-rl-analysis} shows a sharp early rise in oracle best@\(N\), followed by diminishing but nonzero gains over later rollouts, which indicates that the pretrained policy already assigns some probability mass to correct answers but does not concentrate it reliably enough for direct use. This quantity is therefore an oracle support diagnostic, not a deployable accuracy estimate and not an inference-time selector. By contrast, majority@\(N\) remains nearly flat around its single-sample level and does not materially improve with additional samples, showing that naive multi-sample voting fails to recover the latent support revealed by best@\(N\). The format curves rule out parsing as the main bottleneck: the mean parse rate stays close to 1 and the any-parse rate is essentially saturated throughout, so most failures reflect biological decision errors rather than missing terminal labels.

The class-conditional curves further clarify where this support is uneven. \(\mathrm{Up}\) is recovered most easily and approaches saturation after only a small number of rollouts, \(\mathrm{Down}\) improves more gradually, and \(\mathrm{No}\) remains the hardest class across the entire range of \(N\). This separation matters for both DE and DIR: correctly identifying \(\mathrm{No}\) is the main challenge for differential-expression detection, while the persistent gap between \(\mathrm{Up}\) and \(\mathrm{Down}\) shows that direction prediction still requires genuine discrimination between the two changed-state labels. Together, these trends motivate policy optimization that increases the concentration of correct outcomes, rather than relying on more sampling or majority voting at inference time.

\begin{figure}[!h]
\begin{center}
\includegraphics[width=\linewidth]{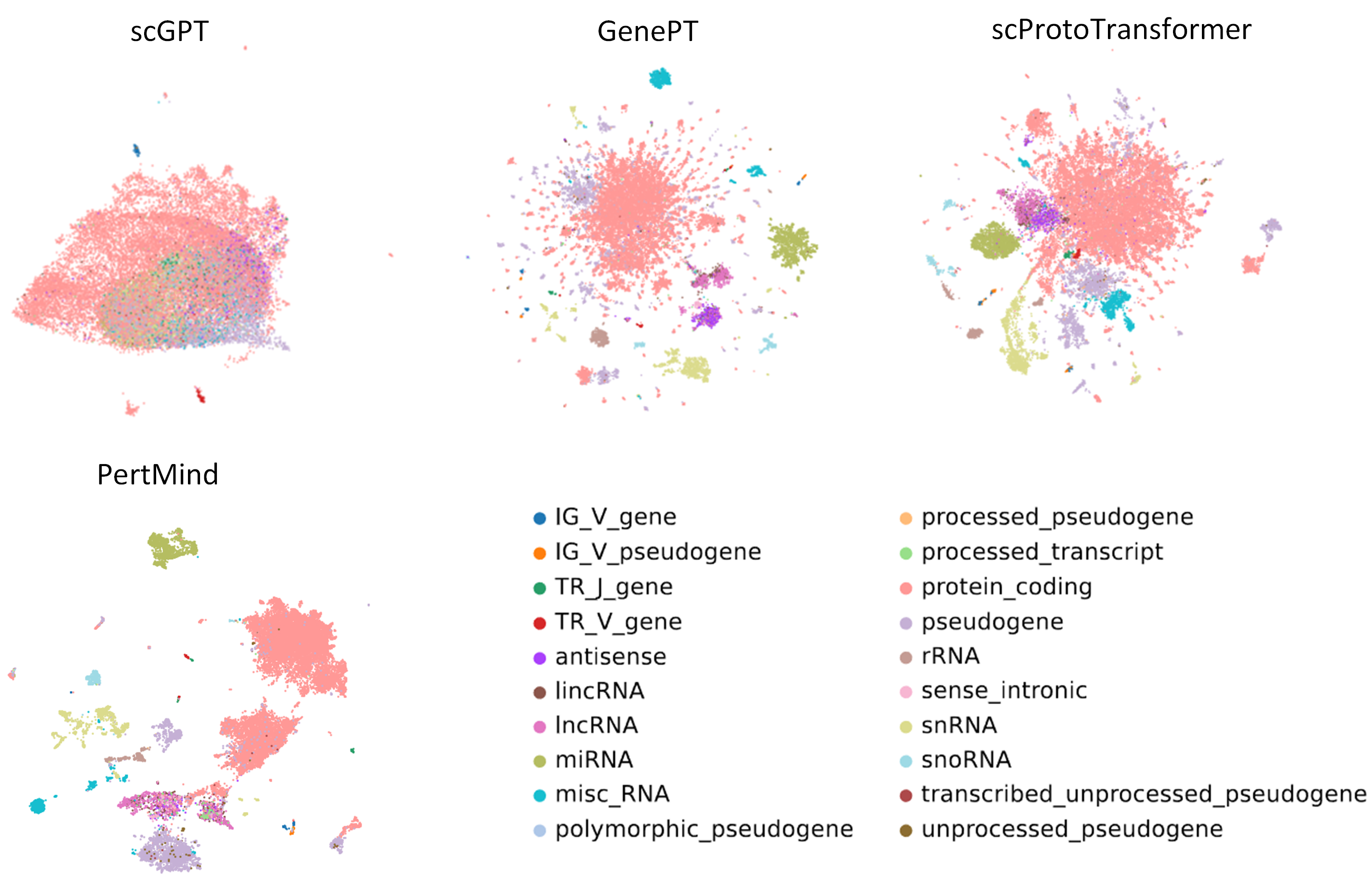}
\end{center}
\caption{Genome-wide UMAP visualization of gene embeddings from scGPT, GenePT, scProtoTransformer, and PertMind. Points denote genes and colors denote annotated gene biotypes. The PertMind map retains distinct biotype-enriched regions, providing a qualitative view of molecular organization complementary to the reference-mapping results in Figure~\ref{fig:multiscale-representation}a. UMAP geometry is sensitive to preprocessing and visualization hyperparameters and is not used as a quantitative performance metric.}
\label{fig:gene-embedding-umap}
\end{figure}

\begin{figure}[!h]
\begin{center}
\includegraphics[width=\linewidth]{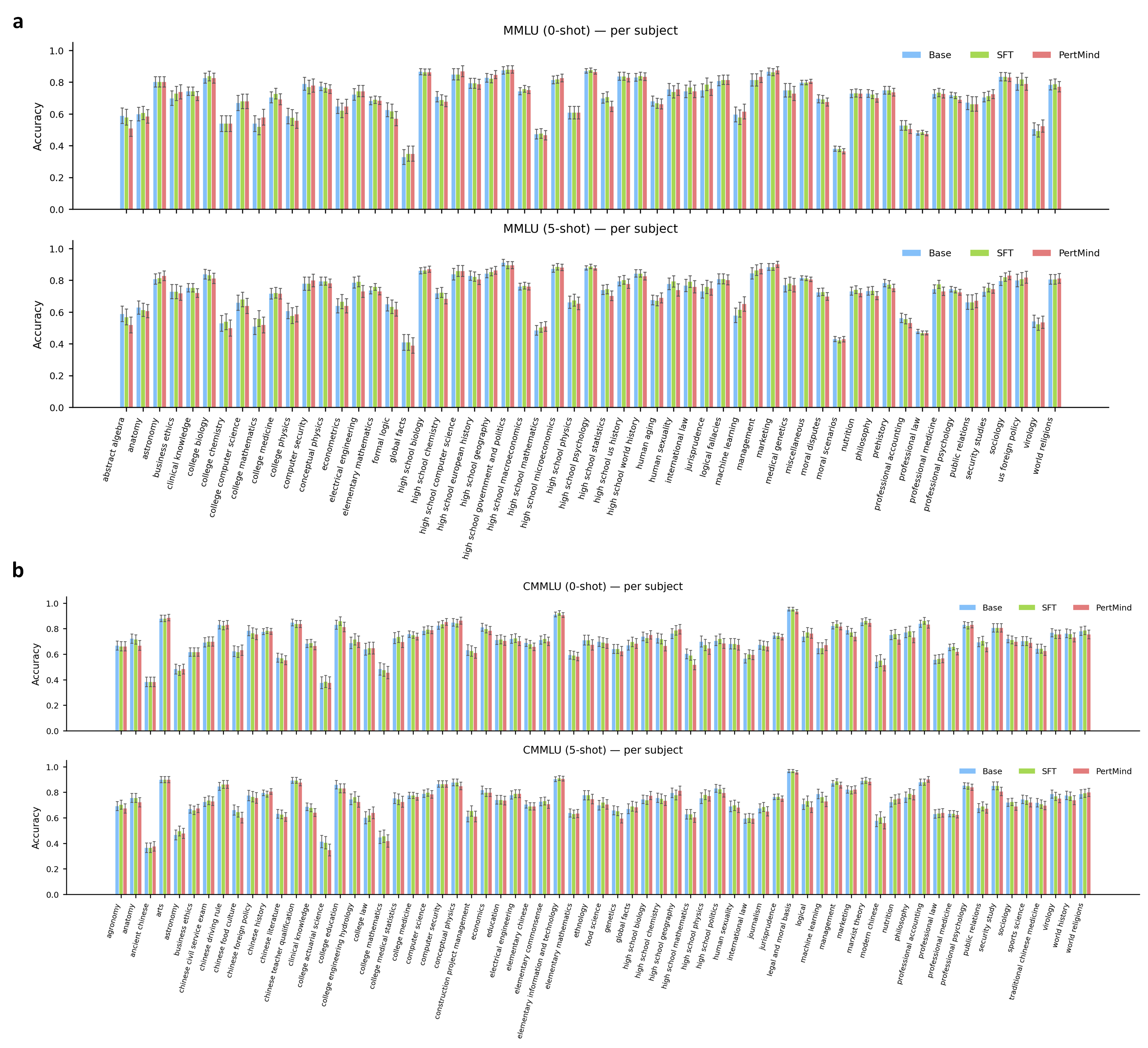}
\end{center}
\caption{Per-subject general-capability evaluation. \textbf{(a)} MMLU accuracy by subject under 0-shot and 5-shot evaluation for Base, SFT, and PertMind. \textbf{(b)} CMMLU accuracy by subject under the same 0-shot and 5-shot settings. Error bars denote standard deviation over five independent runs.}
\label{fig:general-capability}
\end{figure}

\begin{figure}[!h]
\begin{center}
\includegraphics[width=\linewidth]{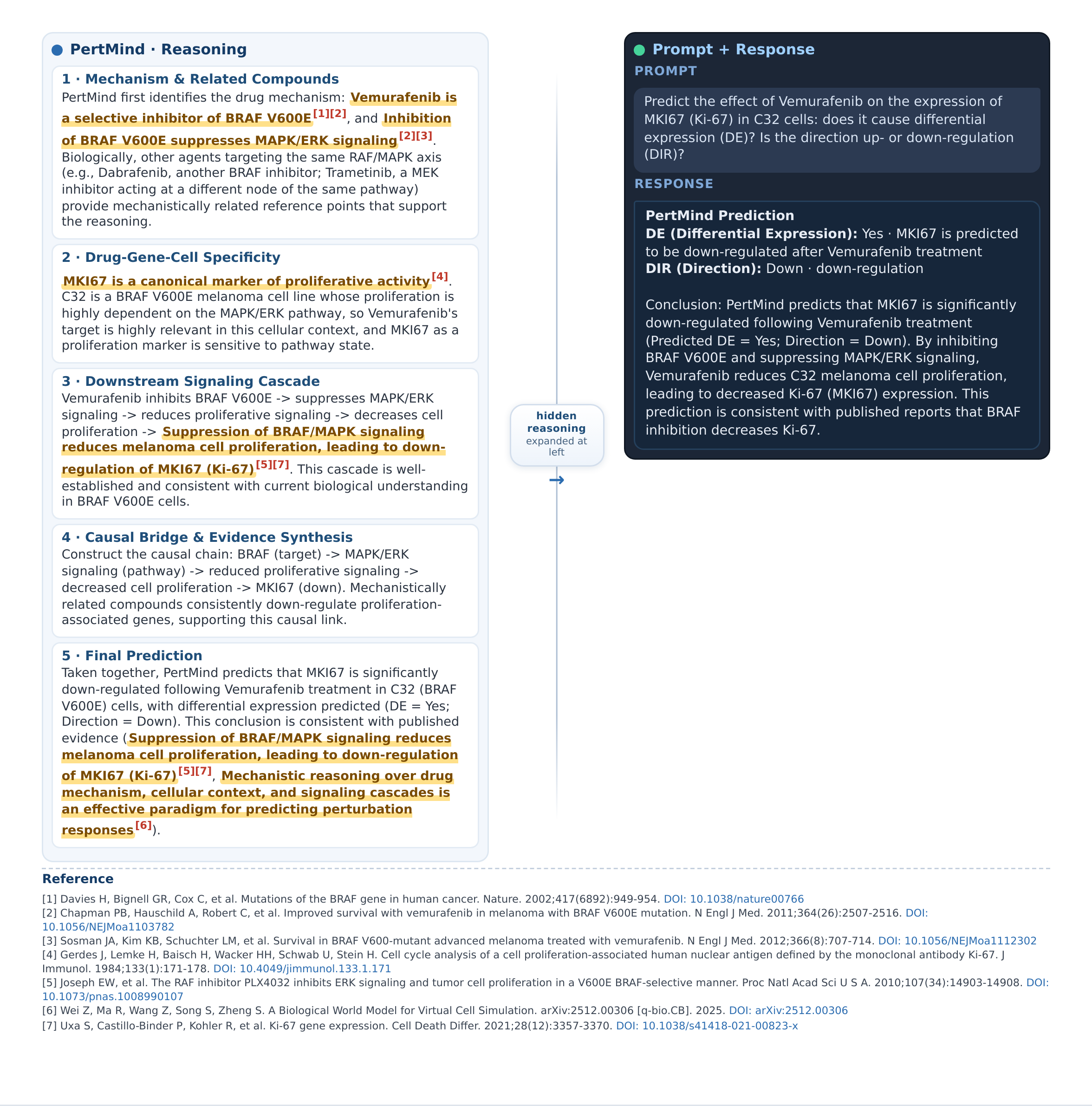}
\end{center}
\caption{Literature-audited free-form natural-language Vemurafenib--C32--MKI67 case. The left side expands the mechanism audit and maps highlighted claims to the references shown in the figure; the right side shows the user prompt and PertMind response. Public literature support for highlighted claims does not establish that the generated reasoning process was faithful.}
\label{fig:natural-language-case}
\end{figure}

\end{document}